\documentclass{article}

\usepackage{arxiv}

\usepackage[utf8]{inputenc} 
\usepackage[T1]{fontenc}    
\usepackage{hyperref}       
\usepackage{url}            
\usepackage{booktabs}       
\usepackage{amsfonts}       
\usepackage{nicefrac}       
\usepackage{microtype}      
\usepackage{lipsum}		
\usepackage{graphicx}
\usepackage{natbib}
\usepackage{newtxtext,newtxmath,algorithm,algpseudocode}
\usepackage{xurl}
\usepackage{subfigure}
\usepackage{doi}
\usepackage{multirow}

\title{Causal-driven attribution (CDA): Estimating channel influence without user-level data}

\author{{Georgios Filippou}\thanks{Corresponding author: \href{email:email-id.com}{filippog@tcd.ie}} \\
	School of Computer Science and Statistics\\
	Trinity College Dublin\\
	Dublin, Ireland \\
	\texttt{filippog@tcd.ie} \\
	\And
	Boi Mai Quach \\
	School of Computer Science and Statistics\\
	Trinity College Dublin\\
	Dublin, Ireland \\
	\texttt{quachm@tcd.ie} \\
	\AND
	Diana Lenghel \\
	School of Computer Science and Statistics\\
	Trinity College Dublin\\
	Dublin, Ireland \\
	\texttt{dlenghel@tcd.ie} \\
	\And
	Arthur White \\
	School of Computer Science and Statistics\\
	Trinity College Dublin\\
	Dublin, Ireland \\
	\texttt{arwhite@tcd.ie} \\
    \AND
	Ashish Kumar Jha \\
	Trinity Business School\\
	Trinity College Dublin\\
	Dublin, Ireland \\
	\texttt{akjha@tcd.ie}
}

\date{}

\renewcommand{\headeright}{Georgios Filippou}
\renewcommand{\undertitle}{}
\renewcommand{\shorttitle}{Causal-Driven Attribution}

\hypersetup{
pdftitle={Causal-driven attribution (CDA): Estimating channel influence without user-level data},
pdfsubject={q-bio.NC, q-bio.QM},
pdfauthor={Georgios Filippou, Elias D.~Striatum},
pdfkeywords={First keyword, Second keyword, More},
}

\begin{document}
\maketitle
Attribution modelling lies at the heart of marketing effectiveness, yet most existing approaches depend on user-level path data, which are increasingly inaccessible due to privacy regulations and platform restrictions. This paper introduces a Causal-Driven Attribution (CDA) framework that infers channel influence using only aggregated impression-level data, avoiding any reliance on user identifiers or click-path tracking. CDA integrates temporal causal discovery (using PCMCI) with causal effect estimation via a Structural Causal Model to recover directional channel relationships and quantify their contributions to conversions. Using large-scale synthetic data designed to replicate real marketing dynamics, we show that CDA achieves an average relative RMSE of 9.50\% when given the true causal graph, and 24.23\% when using the predicted graph, demonstrating strong accuracy under correct structure and meaningful signal recovery even under structural uncertainty. CDA captures cross-channel interdependencies while providing interpretable, privacy-preserving attribution insights, offering a scalable and future-proof alternative to traditional path-based models.


\section{Introduction}
\begin{quote}
``When budgets are tight, marketers need to justify every penny of their spend. But without full-funnel attribution, spending decisions are often based on what is easy to measure rather than what is actually working.'' \\
\textit{--- MarketingWeek, 2025}\footnote{\url{https://www.marketingweek.com/full-funnel-attribution/}}
\end{quote}

Since consumers now learn and respond to marketing stimuli in a multi-channel environment, more sophisticated frameworks are required to evaluate whether investments across these channels actually generate the desired outcomes \citep{berman2018beyond}. This motivation gives rise to the concept of marketing effectiveness, which represents the degree to which marketing investments drive the desired results \cite{solcansky2010measurement}. 

However, measuring marketing effectiveness has become increasingly challenging in the digital era. Many researchers have employed diverse methodologies to evaluate it, including experimentation, qualitative and quantitative observational analyses, econometric modelling, and machine learning approaches. Among these, statistical models have become particularly prominent as rigorous tools for assessing marketing effectiveness \citep{zenetti2014search,han2023cost,labrecque2024native}. Organizations increasingly rely on such models to gain actionable insights guiding strategic decisions, particularly regarding advertising budget allocation under uncertainty \citep{qin2024data}. 

One of the most widely used statistical approaches for defining marketing effectiveness is marketing attribution \cite{buhalis2021bridging}. It has received growing attention as both a methodological and theoretical challenge. Attribution refers to assigning credit to different marketing touchpoints influencing customer purchase paths. Drawing from attribution theory in psychology \cite{kelley1967attribution,heider2013psychology}, this process involves determining causal relationships between marketing stimuli and consumer responses. 
 
The proliferation of entertainment and advertising platforms has added further complexity to understanding consumer behaviour \cite{schweidel2022consumer}. Simultaneously, changes in digital privacy, particularly Apple's iOS14 updates \cite{sokol2021harming}, the European Union’s General Data Protection Regulation (GDPR) \cite{li2019impact}, and the California Consumer Privacy Act (CCPA) \cite{bonta2022california}, have significantly restricted access to individual level data. In a cross-channel interaction environment, where users can move freely across touchpoints and marketing messages appear across numerous platforms and devices, these updates create significant difficulties for accurate measurement due to the lack of data transparency throughout the consumer journey \cite{cui2021informational}.

Limited data availability creates what we term the ``privacy-attribution paradox'' where the need for sophisticated attribution increases as data accessibility decreases. This paradox is compounded by both technical and conceptual limitations that existing literature treats separately but are fundamentally interconnected. Technical limitations include privacy restrictions, limited data access, and systematic mislabelling of traffic sources \cite{starov2018betrayed,filippou2024establishing}. Conceptual limitations arise from the model’s inability to provide transparent interpretability. The results often fail to reveal the precise insights researchers seek. For example, we may desire to understand the causal relationships between marketing channels rather than the correlations among them.

Furthermore, the emergence of view-first (or impression-first) entertainment platforms exemplifies these interconnected challenges. \citet{reiley2012ad} found that standard click-based
models can misattribute the advertising impact by assigning credit to clicks from users who were likely to convert regardless of exposure to the advertising. Users engage deeply with content without necessarily clicking or transitioning to websites, creating influential but unmeasured interactions. Academic research demonstrates that impressions without website visits can generate sales \cite{ghose2016toward}, while industry studies show significant impact of views and impressions on conversion paths without clicks (TikTok for Business 2024\footnote{\url{https://ads.tiktok.com/business/en/blog/beyond-click-attribution-views-impact-purchase }}; Snapchat for Business 2024\footnote{\url{https://forbusiness.snapchat.com/blog/evolving-measurement-neustar}}).

Marketing practitioners are not able to analyse the full user journey since each ad platform tracks its own touchpoints. This creates structural opacity in digital attribution \cite{romero2020digital} which could lead to biased or incomplete attribution outcomes. Thus, the analytical focus shifts from tracing individual paths to inferring the mechanisms that generate them. This is where causal inference becomes essential \cite{ghose2016toward}. In the context of marketing effectiveness, extracting valuable knowledge from the available data continues to be a challenge for both scholars and practitioners. When cause and effect relationships are discovered, support the decision making process of budget allocation \cite{zhou2023direct}. 

Building on these challenges, our study offers two primary contributions. First, we present a theoretical approach that integrates Structural Equation Modelling (SEM) into a temporal causal discovery framework for simulating marketing attribution. Second, we introduces a novel causal-driven attribution (CDA) framework that works entirely on aggregated impression-level data, solving the privacy and data-access limitations of traditional user-level models. To the best of our knowledge, none of the existing models can fully reveal the causal relationships among touch points, including determining the direction of influence between channels and quantifying their impact values while using only aggregated data.

The rest of the paper unfolds as follows. We begin with a review of the relevant literature, grounding our work in the methodological foundations of attribution modelling. We then outline our data simulation process and walk through the modelling approach in detail. This is followed by the presentation of key findings and an extensive model validation process. Finally, we close with a critical discussion of the results and what they mean for both future research and applied marketing practice.

\section{Literature review}
\subsection{Marketing effectiveness and consumer journey}
Marketing effectiveness captures the extent to which marketing activities generate desired business outcomes such as sales, revenue growth, customer acquisition, or brand lift \cite{solcansky2010measurement,raman2012optimal}. Although conceptually simple, evaluating effectiveness becomes increasingly complex once we account for how consumers behave and how marketers attempt to influence that behaviour. 

Classical theories framed the consumer journey as a linear progression from awareness to consideration to purchase \cite{singh2024enhancing}, a structure that made sense in a world with limited media channels. However, these linear frameworks struggle to explain behaviour in contemporary ecosystems where consumers discover, evaluate, and engage with brands across search engines, social platforms, video environments, review sites, apps, and offline stimuli \cite{nam2025navigating}.

The cognitive foundations of such behaviour are well established. Consumers form knowledge structures through repeated exposure \cite{bettman1970information}, rely on associative networks to store and retrieve brand information \cite{alba1987dimensions}, and interpret marketing messages through attributional reasoning \cite{heider2013psychology,kelley1967attribution}.

Digitalisation accelerated these processes by expanding the number of available touchpoints and enabling multi-platform navigation. \citet{verhoef2015multi} demonstrate that how omni-channel environments blur the boundaries between physical and digital experiences, while recent work shows that consumers routinely search on Google, watch YouTube videos, browse Instagram, and ultimately purchase offline \cite{santos2024consumer}. Understanding this fragmented and recursive journey has therefore become central to enhancing marketing effectiveness.

From a managerial perspective, resource allocation theory further complicates the challenge. Behavioural theories emphasise that managers operate under bounded rationality, seeking solutions that are ``good enough'' rather than fully optimal \cite{simon1955behavioral}. As a result, organisations must make decisions under conditions of information overload, internal negotiation, and limited analytical capacity \cite{march1993organizations}.

\citet{cyert2020behavioral} argue that firms do not optimise in the textbook sense. Instead, they rely on routines, satisficing, and sequential budget adjustments shaped by internal bargaining. As media channels proliferate and privacy constraints reduce the availability of granular data, these behavioural limitations become even more pronounced. Recent work highlights the deep uncertainty marketing leaders face when evaluating investment effectiveness, thereby increasing demand for reliable and transparent measurement frameworks \cite{malter2020past}.

Within this context of behavioural complexity, fragmented data, and managerial uncertainty, attribution modelling has emerged as a central tool for accountability and optimisation.

\subsection{Marketing Attribution}
\subsubsection{The heuristic attribution era}
The early days of marketing attribution relied on pre-defined rules that assigned credit to specific marketing touchpoints \cite{donndelinger2020design}. Google’s attribution framework, developed between 2006 and 2015, introduced heuristic models such as first-interaction, linear, time-decay, position-based, and last-click \cite{gaur2020attribution,gaur2024maximizing}. These rule-based models quickly became popular among practitioners and researchers in online marketing, and many of them remain in use today.

Out of the various options and models, last-click, which assigns 100\% of the credit to the final touchpoint before conversion, gained significant traction as the main choice for both businesses and media companies. Despite its conceptual weaknesses, last-click became the industry default. However, the adoption of last-click attribution increased even as criticism mounted \cite{kufile2023developing}. This persistence reflects organisational behaviour: under uncertainty, firms gravitate toward simple and interpretable rules, even when these rules are inaccurate or questionable.

With the rise of digital media over the past decade, consumers have more options to engage with brands across multiple platforms. Likewise, brands are able to distribute their content and advertising message in more platforms \cite{dens2023rise}. As a result heuristic models struggled to provide sufficient information around the channel which was solely responsible for key business activities such as purchases \cite{tao2025graphical}. 

First, it  reflects reductive assumptions about consumer behaviour and fails to capture the full influence of earlier marketing interactions \cite{el2024evolution}. Second, researchers highlight the increasingly non-linear nature of consumer journeys \cite{santos2024consumer}. Contemporary consumer behaviour theory recognises highly fragmented journeys in which users move across multiple devices and platforms, interacting with both organic and paid content before making purchase decisions \cite{nam2025navigating}.

To mitigate these challenges, researchers came up with a different approach: data-driven models \cite{romero2020digital}. For some, data-driven models are considered as ``next generation attribution models'' because they are grounded in mathematical and statistical methods rather than pre-defined rules.

\subsubsection{The data-driven attribution era}

Consumers now engage with brands across multiple platforms and devices \cite{berman2018beyond,santos2024consumer,nam2025navigating}, while brands have simultaneously expanded their presence across an increasing number of digital touchpoints \cite{dens2023rise}. This behavioural shift created the need to analyse entire paths rather than isolated events. Consequently, data-driven attribution models emerged as a methodological evolution, designed to process path data, meaning the full sequence of user touchpoints before conversion, and thereby capture the complexity of modern consumer journeys \cite{hui2009path,kannan2016path}. Several data-driven methods currently support marketing attribution decisions. Below, we highlight the approaches that have shaped the industry and discuss recent developments and emerging trends.

Shapley values, borrowed from cooperative game theory, provide an elegant mathematical framework for distributing credit across marketing touchpoints \cite{shapley1953value}. Early work demonstrated the versatility of influence-attribution methodologies, which opened the door to marketing applications \cite{papapetrou2011shapley}. The method was first pioneered for marketing use in work that positioned it for practical application while acknowledging the limitations of observational data \cite{dalessandro2012causally}. The approach is theoretically sound: it calculates each touchpoint’s marginal contribution across all possible combinations and then assigns proportional credit. Later research improved computational efficiency, incorporated channel synergies, and expanded the method’s applicability \cite{yadagiri2015non,nisar2015purchase,zhao2018shapley}. Further developments also introduced hybrid approaches that combine Shapley values with complementary modelling techniques \cite{kadyrov2019attribution}.

Markov chains offered an alternative framework by modelling consumer journeys as probabilistic state transitions \cite{pfeifer2000modeling,homburg2009managing}. The ``adgraph'' concept was later introduced to represent journeys as graphs in which nodes capture ad events and edges encode transition probabilities \cite{archak2010mining}. This approach proved effective for identifying path dependencies and drop-off points, a conclusion supported by comprehensive empirical validation comparing simple, forward, and backward Markov chains \cite{anderl2014mapping}. Subsequent extensions further demonstrated its applicability across industries \cite{kakalejvcik2018multichannel,sandeep2021channel}.

The integration of machine learning (ML) into attribution modelling provides a rigorous, data-driven approach capable of detecting patterns in customer behaviour and assessing the marginal influence of each touchpoint. For instance, \citet{shao2011data} introduced bagged logistic regression for attribution. It is worth mentioning, their work was the first published use of the term ``data-driven attribution.'' Subsequent studies have developed models that distinguish between short- and long-term effects across cohorts \cite{breuer2012short}, as well as parametric and non-parametric frameworks that account for spillover effects \cite{li2014attributing}. Amazon’s MTA approach leverages both randomized controlled trials (RCTs) and machine learning models to determine how conversion credit should be distributed across Amazon Ads touchpoints \cite{lewis2025amazon}. Other advances apply survival analysis to capture time-decay dynamics \cite{zhang2014multi} and utilise hazard-rate modelling to quantify the influence of individual ad exposures \cite{ji2017additional}.

Building on these ML-based advances, deep learning methods have recently been applied to the MTA problem, offering the ability to model nonlinear interactions, temporal dependencies, and user heterogeneity at scale. The Deep Neural Net with Attention (DNAMTA) model integrates attention mechanisms to account for temporal effects and user characteristics, achieving strong predictive performance while producing interpretable touchpoint-level contribution estimates \cite{arava2018deep}. The Dual-Attention Recurrent Neural Network (DARNN) further advances this idea by learning attribution weights directly from the conversion objective through a dual-attention architecture, dynamically allocating credit across a user’s full exposure sequence \cite{ren2018learning}. Complementing these approaches, DeepMTA combines phased-LSTM networks with an additive attribution method based on Shapley values, delivering both accurate conversion prediction and transparent explanations of each touchpoint’s marginal effect \cite{yang2020interpretable}. Collectively, these deep learning frameworks demonstrate how neural models can capture richer temporal structure and provide more interpretable attribution estimates than traditional methods.

\subsection{The causal imperative}
\label{subsec:causal_imperative}
Imagine a marketer running a sale campaign across several channels, including email, social, and display. Historical data may show that customers who receive more email messages tend to convert at higher rates. However, choosing the most effective channels remains challenging because correlations do not reveal the true incremental effect of each touchpoint. For instance, even if higher email volume coincides with higher conversions, it would be misleading to attribute all of those conversions to email. Many of these users may have already formed their purchase intent through other channels, such as search or social media, long before the email was delivered. Therefore, correlation-based attribution models can misallocate credit across channels because they are incapable of distinguishing true causal effects from correlations.

Causal inference and marketing effectiveness have become increasingly intertwined \cite{hair2021data}, a shift fuelled by advances in data availability, experimental design, and machine learning. The emphasis on causality has moved the field beyond correlational attribution toward frameworks aimed at identifying true cause–and-effect relationships. Conceptually, this transition reflects a broader recognition that understanding why marketing works is essential for credible measurement and strategic decision-making \cite{zhou2023direct}.

Causal-based approaches to multitouch attribution address the limitations of correlational models by estimating the incremental effect of each advertising touchpoint. \citet{dalessandro2012causally} frame attribution as a causal estimation problem, using Shapley values to allocate credit based on each ad’s marginal effect. \citet{kumar2020camta} introduce CAMTA, a causal recurrent neural network model designed to account for time-varying confounders and reduce selection bias, although its use of click behaviour as pseudo-feedback may introduce additional sources of bias. \citet{yao2022causalmta} present CausalMTA, which aims to eliminate confounding bias from both static user attributes and dynamic behavioural features through counterfactual prediction. \citet{tang2024dcrmta} advances this line of research with DCRMTA, a deep causal representation model that isolates causal features linking user behaviour to conversion while reducing the influence of confounding variables.

Different from the individual-level data used in most studies, \citet{kireyev2016display} utilise a dataset consisting only of aggregate consumer behaviour to examine the dynamic interaction between paid search and display advertising. They combine time-series analysis with Granger causality to identify cross-channel spillovers and broader advertising dynamics. However, Granger causality does not constitute true causal inference \cite{shojaie2022granger}. It shows only that past values of one time series improve the prediction of another, not that one variable exerts a causal effect from the traditional causal-inference sense.

\subsection{Structural challenges in attribution}
\label{subsec:regulation_challenges}
Most established attribution techniques, including Shapley value models, Markov chains, deep learning architectures, and even recent causal MTA frameworks, depend on granular, event-level behavioural data to reconstruct user paths. However, in a privacy-first ecosystem shaped by regulations such as GDPR \cite{li2019impact}, CCPA \cite{bonta2022california}, and platform-level changes such as Apple’s iOS14 \cite{sokol2021harming}, access to persistent identifiers has been severely restricted. As a result, the data required to support these models has become increasingly unavailable, inconsistent, or systematically mislabelled \cite{starov2018betrayed}.

Furthermore, a growing share of digital engagement now occurs in channels that do not produce click-based signals. The industry has been massively shaped by short-form video, influencers, connected TV, and impression-first entertainment platforms \cite{yuan2022effect}. These environments generate meaningful behavioural impact through exposure rather than direct interaction. \citet{ghose2016toward} indicate that impressions without website visits can materially influence conversions. Evidence from \citet{filippou2024establishing} further demonstrates that media exposure can influence outcomes indirectly, for example by increasing organic branded searches that do not appear in click sequences. Traditional path-based attribution, which relies on sequences of clicks or identifiable transitions, cannot capture these impression-driven effects.

The marketing productivity literature consistently argues that marketing investments should be evaluated based on their causal contribution to business outcomes \cite{sheth2002marketing,morgan2002marketing,rust2004measuring}. Correlation-based attribution models, whether heuristic or data-driven, are prone to biased estimates, particularly in environments characterized by endogenous targeting, cross-channel spillovers, and unobserved confounders. In the absence of causal evidence, budget allocations can become systematically distorted, limiting firms’ ability to assess true ROI. 

\subsection{Towards a new paradigm}
Industry practice and academic literature increasingly converge on the same conclusion: today’s environment exposes a critical measurement gap. Existing attribution frameworks break down when only aggregated or impression-level data are available, particularly for smaller or emerging channels that do not provide persistent user identifiers \cite{cui2021informational,ghose2016toward}. They also fail when user journeys are unobservable due to regulatory changes, when touchpoints do not generate clicks as in impression-first platforms, and when privacy constraints prohibit the collection of path-level behavioural data (as discussed in Section \ref{subsec:regulation_challenges}).

While a business could build complex API integrations across multiple platforms to capture user-level path data, it must carefully weigh the trade-off between adopting a new solution and maintaining the existing one at a high and potentially unsustainable infrastructure cost. Even with sufficient data, certain traditional MTA and deep learning models can be difficult to deploy due to their heavy computational requirements. \citet{singh2024enhancing} emphasise that computational efficiency is essential for modern customer-journey analytics, and models that require excessive computing resources are unlikely to be adopted in practice.

It is evident that there is a need for a new family of attribution methodologies. The shift from individual to aggregate data, from correlation to causation, and from clicks to impressions reflects more than a methodological evolution. It represents a fundamental reconceptualization of marketing measurement, one that aligns with how consumers behave and how privacy-conscious societies choose to govern their data. Our framework offers one step toward this reconceptualization.

Our work advances this direction through a causal-driven attribution (CDA) methodology that:

\begin{enumerate}
    \item Operates entirely on impression-level, aggregated time-series data.
    \item Explicitly designed for privacy-restricted environments.
    \item Accommodates channels that do not generate user-click signals.
    \item Identifies both direct and indirect causal relationships across marketing touchpoints.
    \item Remains computationally lightweight and accessible to advertisers without complex data engineering requirements.
\end{enumerate}

\section{Methodology}
\subsection{Data Generation}
\label{subsec:data-generation}
Data accessibility for advertisers has been significantly affected by both regulatory and industry changes \cite{wernerfelt2025estimating}. For instance, accessing user-level data typically requires API access (Facebook, 2025)\footnote{\url{https://developers.facebook.com/docs/graph-api/overview}}. However, using marketing APIs effectively also demands additional infrastructure, which is often available only to larger corporations \cite{simon2021apis}. Additionally, \citet{cui2021informational} clearly outlined the challenges associated with multi-touch attribution and how data limitations hinder its application. 

Initially, we aimed to obtain a dataset that would enable a comparison between traditional multi-touch attribution methods and our proposed approach. However, after nearly two years of exploration, we were unable to find a suitable dataset. To the best of our knowledge, all existing publications on advertising and causal learning rely on only two publicly available datasets \cite{job2025exploring}, neither of which meets the requirements for this work. Due to the lack of ground truth in real-world causal marketing data, simulation has played a vital role in method development and evaluation \cite{herman2025unitless}.

Motivated by these challenges, we developed an approach that allowed us to generate a synthetic dataset. Our aim is not only to create a comprehensive and representative marketing dataset but moreover to be able to establish the so-called ground truth required for model validation, thereby bypassing the problem of insufficient data. 

We acknowledge that real marketing data cannot be perfectly simulated due to hard-to-model factors such as customer awareness and behaviour are influenced by psychological aspects. Therefore, several simplifying assumptions were made to ensure that our causal-driven attribution model can potentially be applied in real-world scenarios. Based on these assumptions, we incorporated baseline impressions, growth rate, and conversion rate for each channel, as well as random noise to capture variability. Cross-influence of marketing channels introduced as the last step. By incorporating cross-influence, we are able to have channels as confounding variables which is essential for causal inference conversation \cite{suttorp2015graphical}.

\subsubsection{Causal DAG Graph}
Before crunching any numbers, we first need a high-level understanding of the causal relationships among our marketing channels in order to define the scenario for which we will generate data. Correlation only describes statistical associations between variables and cannot answer questions such as “Does changing the value of Channel A lead to a change in Channel B?” To simulate realistic marketing behaviour, we need a visual representation of the underlying causal structure, and this is where Directed Acyclic Graphs (DAGs) come into play.

\citet{pearl2009causality} explains that DAGs provide a natural way to represent causal structures. Causal effects arise from underlying, unobserved processes that can be viewed as the equilibrium outcomes of a system of behavioural equations, which themselves form a simplified model of how the world operates. Graph notation captures this structure efficiently: nodes represent random variables that arise from some underlying data-generating process, and arrows indicate causal influence between variables. The direction of each arrow corresponds to the direction of causality. The acyclic property ensures that the graph does not contain feedback loops, which prevents circular causation and preserves clarity in causal interpretation. Additionally, causal effects may be direct (e.g., $X \rightarrow Y$), or indirect (e.g., $X \rightarrow Z \rightarrow Y$), where the influence is transmitted through a mediating variable.

\begin{figure}[!t]
\centering
\includegraphics[width=0.55\textwidth,height=0.40\textwidth]{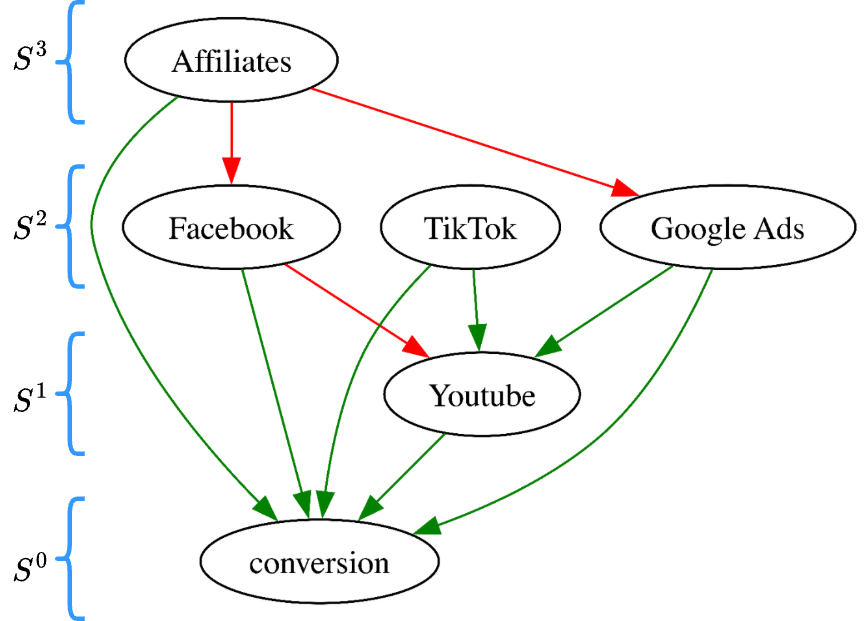}
\caption{Example of a DAG's structure, including nodes: \textit{Affiliates}, \textit{Facebook}, \textit{TikTok}, \textit{Google Ads}, \textit{YouTube}, and \textit{conversion}. Edges represent directed causal influences: green arrows indicate positive values (putative amplification/halo), and red arrows indicate negative values (putative substitution/crowd-out). Nodes are grouped into causal layers as follows:  
$S^3$: \textit{Affiliates},  
$S^2$: \textit{Facebook}, \textit{TikTok}, \textit{Google Ads},  
$S^1$: \textit{YouTube},  
$S^0$: \textit{conversion}.}
\label{fig:causal_DAG_simulation}
\end{figure}

A directed acyclic graph (DAG) $\mathcal{G}$ is a visual structure consisting of a set of nodes (vertices) $\mathbf{V}$ and a set of directed edges (arrows) $\mathcal{E}$. Each node $\mathbf{V}_i$ is associated with a random variable $X_i$. In a DAG, there are no directed paths that begin and end at the same node. Formally, no cycles exist such that a path starts at $\mathbf{V}_i$ and returns to $\mathbf{V}_i$ (either directly or indirectly). A finite set $S$ denotes the causal layers (also known as a non-signalling set), which are subsets of nodes that are causally independent (i.e., nodes within the same layer cannot signal to one another) \cite{giarmatzi2019quantum,nashed2025causal}. Figure~\ref{fig:causal_DAG_simulation} illustrates the structure of a DAG, including its nodes, edges, and layered decomposition.

\begin{figure}[!t]
\centering
\includegraphics[width=0.52\textwidth,height=0.48\textwidth]{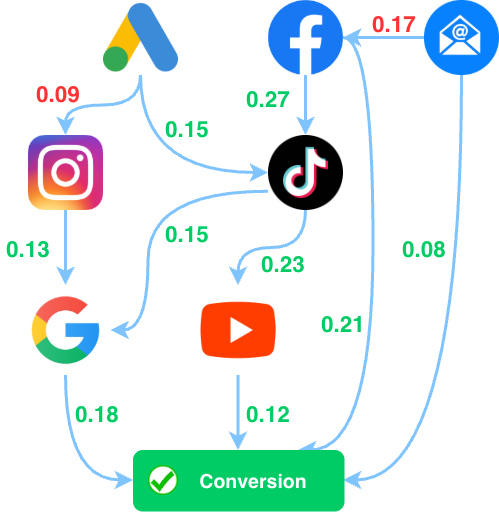}
\caption{Example of a causal DAG showing the structural relationships among marketing channels. Some channels act as upstream drivers that influence conversions indirectly through intermediate touchpoints, while others have a direct impact on the outcome. All effects flow in a single direction with no circular feedback, providing a clear view of how influence moves through the system.}
\label{fig:causal_DAG}
\end{figure}

Recent literature highlights a growing trend toward using DAGs as an uncertainty-aware and interpretable tool for visualizing causal graphs \cite{xie2020visual}. DAGs provide a transparent way to distinguish causal influence from correlation. In marketing attribution, they map the structural relationships among channels and outcomes, clarifying how touchpoints interact, mediate, or confound one another throughout the consumer journey. Figure~\ref{fig:causal_DAG} illustrates an example of causal DAGs in a marketing context and shows how channels interact and ultimately influence conversions. According to the generated DAGs, we can build mathematical equations based on the following part to run the simulation.

\subsubsection{Assumptions and Equations}
\label{subsubsec:assumptions_equations}
We acknowledge that real marketing data cannot be perfectly simulated due to hard-to-model factors such as customer awareness and behaviour are influenced by psychological aspects. Additionally, since causal mechanisms in real-world marketing systems are known with high uncertainty \cite{runge2019inferring,brouillard2025landscape}, the simulation must reflect the properties of the real datasets on which the algorithms will be applied. This is necessary to ensure that our causal-driven attribution developed using simulated data can potentially translate to real-world scenarios.

A standard linear equation has been widely used to model the activity changes of each channel \cite{danaher2018delusion,zhao2019revenue}. Suppose $x_t$ represents the number of impressions for channel $X = (x_1, x_2 ... x_t, x_{t+1}...)$ at time $t$; the linear trend is defined as follows:

\begin{equation}
    x_t = \alpha + \beta t + \epsilon_t,
\end{equation}

where $x_t$ is the channel value at time $t$, $\alpha$ represents the baseline impressions, $\beta$ is the growth rate of the channel, and $\epsilon_t$ denotes the random noise.

The next step was to establish the cross-influence between channel coefficients. 
Taking two channels as an example, let $X_1$ represent \textit{Search advertising}\footnote{\textbf{Search advertising:} We refer to Search advertising (Search Ads) as a form of paid digital marketing in which advertisers bid to display sponsored messages alongside organic search results on search engines such as Google or Bing.} 
and $X_2$ denote \textit{Social advertising}\footnote{\textbf{Social advertising:} We refer to Social advertising (Social Ads) as paid digital communication delivered through social networking platforms such as Facebook, Instagram, TikTok, or LinkedIn. It allows advertisers to target audiences based on demographic, psychographic, and behavioral data.}. 
Each channel still follows a linear model; however, they now influence each other directly within the same time period $t$ (i.e., no lag), as follows:

\begin{equation}
\begin{aligned}
    x_{1,t} &= \alpha_1 + \beta_1 t + w_{12} x_{2,t} + \epsilon_{1,t}, \\
    x_{2,t} &= \alpha_2 + \beta_2 t + w_{21} x_{1,t} + \epsilon_{2,t},
\end{aligned}
\end{equation}

where $x_{1,t}$ and $x_{2,t}$ represent the impression amounts of 
\textit{Social Ads} and \textit{Search Ads} at time $t$, respectively; 
$\alpha_1$ and $\alpha_2$ denote the baseline impressions for each channel; 
$\beta_1$ and $\beta_2$ are their respective growth rates; $\epsilon_{1,t}$ and $\epsilon_{2,t}$ represent random noise terms. 

More importantly, \textbf{\boldmath $w_{12}$} measures how much \textit{Social Ads} influence \textit{Search Ads}, 
and \textbf{\boldmath $w_{21}$} measures how much \textit{Search Ads} affect \textit{Social Ads}. These terms capture the directional causal effects between channels. Without them, the relationship would represent only correlation rather than causation.

Extending the idea to multiple channels, each channel’s activity is determined not only by its own trend but also by the concurrent activity levels of the other channels (no lag):

\begin{equation}
x_{i,t} = \alpha_i + \beta_i t + \sum_{j \neq i} w_{ij} x_{j,t} + \epsilon_{i,t},
\end{equation}

where $\alpha_i$ and $\beta_i$ represent the baseline and growth trend of channel $i$, respectively, 
while $w_{ij}$ measures how the current activity of channel $j$ influences channel $i$. 
$\epsilon_{i,t}$ denotes the random noise term for channel $i$.

Next, we transition from impressions to conversions per channel using the following relationship:

\begin{equation}
y_{t,i} = \gamma_0 + \sum_{i=1}^{N} \gamma_i x_{i,t} + \eta_t,
\end{equation}

where $y_{t,i}$ denotes the number of conversions for channel $i$, 
$\gamma_i$ represents the effectiveness of channel $i$, 
and $\eta_t$ is the random noise term. 
This formulation captures the direct, concurrent (no-lag) causal influence among marketing channels.

Lastly, the total number of conversions at each time step $t$ can be expressed as the sum of baseline (non-marketing) conversions and the conversions attributed to all marketing channels.

\begin{equation}
C_t = B_t + \sum_{i=1}^{N} y_{t,i},
\end{equation}

where $C_t$ represents the total conversions at time $t$, $B_t$ is the baseline component capturing conversions not driven by advertising activities (e.g., organic or repeat behavior), 
and $y_{t,i}$ corresponds to the conversions generated by marketing channel $i$ at the same time step.

This simulation integrates both organic and paid contributions to overall performance, allowing the model to disentangle the inherent baseline trend from the incremental impact of each channel. It provides a comprehensive view of how marketing efforts collectively contribute to total conversions while maintaining interpretability at the channel level.

\begin{figure}[!htpb]
\centering
\includegraphics[width=0.90\textwidth]{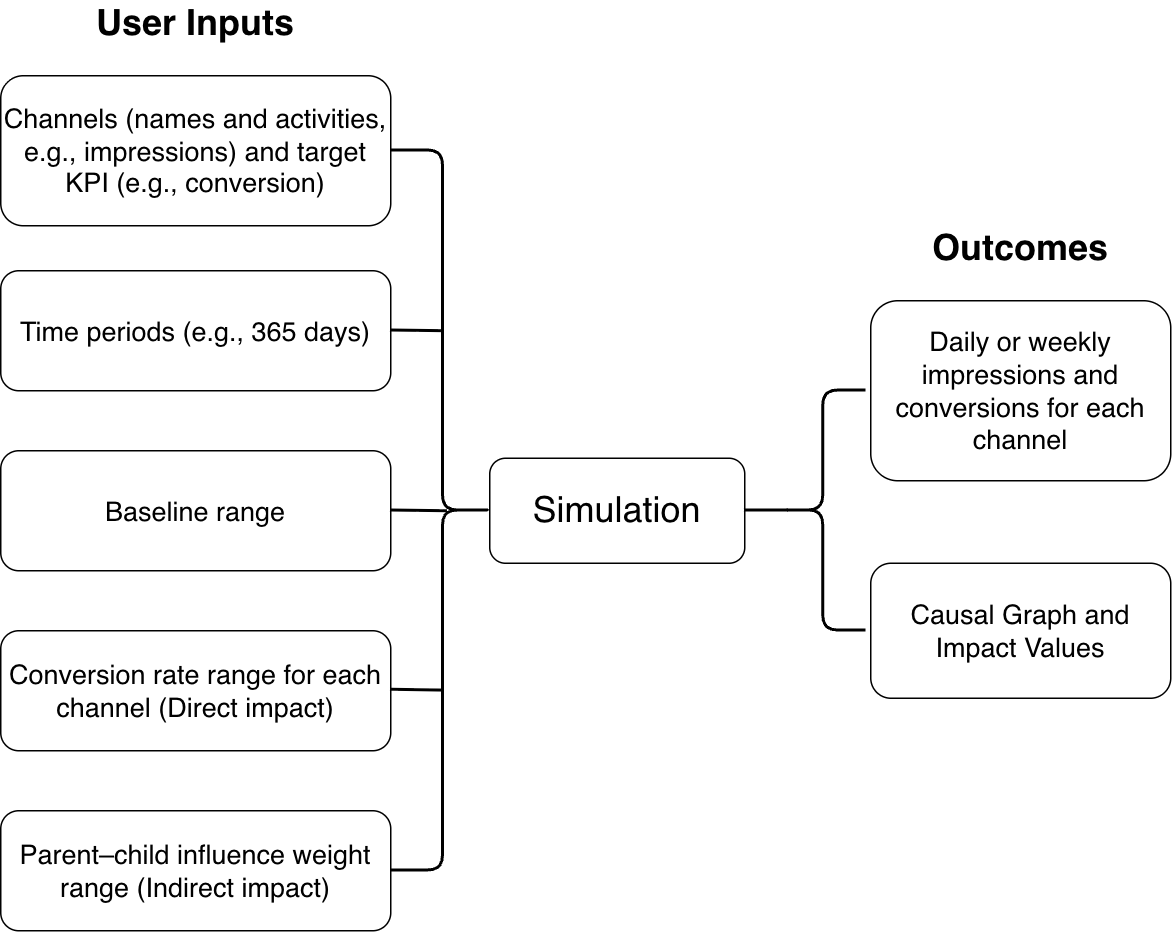}
\caption{Overview of the simulation framework.}
\label{fig:simulation_framework}
\end{figure}

Figure~\ref{fig:simulation_framework} illustrates the structure of the proposed simulation framework for generating synthetic causal marketing attribution data. The process begins with user-defined inputs that specify the simulation context. These include: (1) the set of marketing channels and their associated activities (e.g., impressions) along with the target key performance indicator (KPI) such as conversions; (2) the time period over which the simulation runs (e.g., 365 days); (3) a baseline range representing conversion levels independent of media activity; (4) the conversion-rate range for each channel, which determines the direct impact of media exposure on conversions; and (5) the parent–child influence weight range, capturing indirect effects that arise from causal dependencies among channels.

\begin{figure}[!t]
\centering
\includegraphics[width=0.90\textwidth]{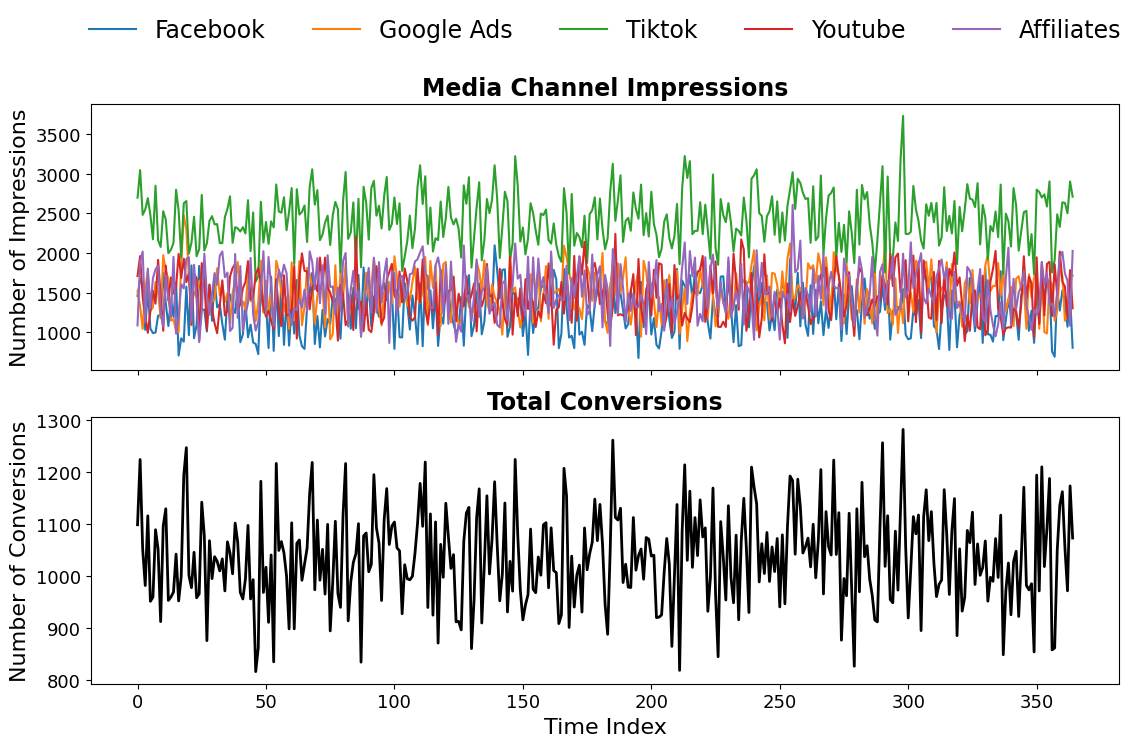}
\caption{A random simulated data illustrating marketing channel impressions and conversions within 365 days. The top panel presents the impression trends for five marketing channels, including Google Ads, Affiliates, TikTok, Facebook, and YouTube. Each series follows a distinct stochastic pattern generated from a linear growth process with random noise and inter-channel influence.}
\label{fig:trend_simulation}
\end{figure}

\begin{figure}[!t]
\centering
\includegraphics[width=0.55\textwidth]{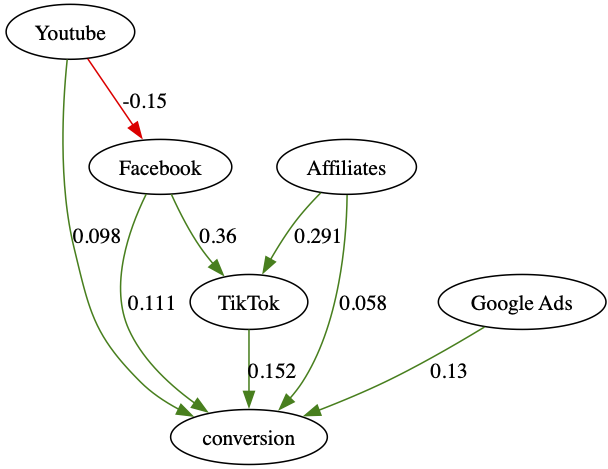}
\caption{Example of a DAG output representing multi-channel interactions from randomly simulated data. Each node represents a marketing channel (Affiliates, Google Ads, TikTok, Facebook, YouTube) or the target outcome (Conversions). Directed arrows denote causal influence between nodes, where the direction of the arrow indicates the flow of information from cause to effect. Green arrows correspond to positive causal weights, indicating that an increase in the originating channel’s activity leads to a subsequent increase in the target channel or conversions. 
}
\label{fig:dag_simulation}
\end{figure}

Given these inputs, the simulation module produces two categories of outputs. The first is the time-series data of daily or weekly impressions and conversions for each channel (as shown in Figure~\ref{fig:trend_simulation}), reflecting both direct and mediated effects. The second is a causal graph with corresponding impact values that quantify the impact values and direction of relationships among channels and between channels and the target KPI (Figure~\ref{fig:dag_simulation}). Together, these outputs enable the analysis of causal influence and attribution in a controlled, reproducible environment.

\subsection{Modelling}
Attribution is fundamentally a cause-and-effect problem. Understanding causality in attribution is a core challenge for any discipline that involves decision-making under uncertainty \cite{kelly2018causal}. A causal model identifies and maps the qualitative cause-and-effect relationships and combines them with quantitative information about the strengths of those relationships. Therefore, our approach uses a two-stage framework: first, we discover the causal relationships among marketing channels; then we estimate the causal influence (weights) of each channel on the target KPI (e.g., conversions).

\subsubsection{Causal DAG Discovery}
Through data simulation, we can obtain the causal relationships between channels and conversions. In reality, however, businesses typically have only time series data representing channel activities and total conversions over time, similar to the simulated multi-channel dataset with the schema shown in Table~\ref{tab:data_schema}. In practice, incrementality has become a common approach to measure causality, as it identifies what actually caused a lift in performance. A causal DAG can be constructed using prior knowledge combined with numerous tests, such as lift tests or A/B experiments. However, this expert-driven, test-and-iterate approach is expensive and time-consuming. Discovering the causal structure that governs the relationships among variables purely from observational data remains a challenging and valuable problem. In this work, we apply PCMCI, an automated procedure for discovering causal graphs based on graphical causal modeling and conditional independence principles. Moreover, PCMCI scales effectively to large time series datasets and accommodates linear, nonlinear, and time-delayed dependencies.


\begin{table*}[!htbp]
\centering
\caption{Schema of the simulated multi-channel marketing dataset.}
\label{tab:data_schema}
\begin{tabular}{lll}
\hline
\textbf{Column} & \textbf{Type} & \textbf{Description} \\ \hline
\texttt{index} & integer & Time step index (e.g., day or week). \\
\texttt{conversion} & float & Total conversions across all marketing channels. \\
\texttt{channel\_1\_impression} & float & Impressions from marketing channel 1. \\
\texttt{channel\_2\_impression} & float & Impressions from marketing channel 2. \\
\texttt{channel\_3\_impression} & float & Impressions from marketing channel 3. \\
\texttt{...} & ... & ... \\
\texttt{channel\_n\_impression} & float & Impressions from marketing channel $n$. \\
\\ \hline
\end{tabular}

\vspace{0.5cm}

\noindent
\raggedright
\textit{Note:} The dataset contains 365 daily observations across $n$ simulated marketing channels. Each row corresponds to a single time step, including the number of impressions per channel and total conversions.

\end{table*}

\paragraph{\textbf{Causal network discovery with PCMCI}}

Consider an underlying multivariate time series $\mathbf{X}_t = \big(X_t^{(1)}, X_t^{(2)}, \ldots, X_t^{(d)}\big)$, where each component represents the daily activity of a marketing channel (e.g., Facebook Ads, TikTok, Google Ads, Affiliates). Our objective is to infer the causal structure governing how past channel activities influence current outcomes. Time lag $\tau$, also know as ``time-delay causality'', refers to the temporal interval between a cause and its effect. We assume finite-order temporal dependencies up to a maximum lag $\tau_{\max}$ and seek, for each variable $X_t^{(i)}$, to estimate the set of causal parents
\begin{equation}
    \mathcal{P}(X_t^{(i)}) \subseteq
    \big\{ X_{t-\tau}^{(j)} \;\big|\; 1 \le j \le d,\; 1 \le \tau \le \tau_{\max} \big\}.
\end{equation}

PCMCI proceeds in two stages. In the first stage, a PC$_1$ condition-selection step identifies relevant 
conditioning sets $\widehat{\mathcal{P}}(X_t^{j})$ for each variable 
$X_t^{j}$. Starting from all lagged variables up to $\tau_{\max}$, variables 
that are conditionally independent of the target given subsets of the remaining 
candidates are removed using a conditional independence test (e.g., partial 
correlation, ParCorr) performed at significance level $\alpha_{\mathrm{PC}}$.

In the second stage, PCMCI applies the Momentary Conditional Independence (MCI) 
test to evaluate whether a lagged variable $X_{t-\tau}^{i}$ is a causal parent 
of $X_t^{j}$. For each variable pair, MCI tests
\begin{equation}
    X^{i}_{t-\tau} \;\not\!\perp\!\!\!\perp\; X^{j}_{t}
    \;\big|\;
    \widehat{\mathcal{P}}(X^{j}_{t}) \setminus \{ X^{i}_{t-\tau} \},
    \; \widehat{\mathcal{P}}(X^{i}_{t-\tau}) ,
\end{equation}
using an appropriate conditional independence test (e.g., ParCorr) at significance
level $\alpha_{\mathrm{MCI}}$. If the null hypothesis of conditional independence 
is rejected, the directed lagged edge $X^{i}_{t-\tau} \to X^{j}_{t}$ is included 
in the estimated causal graph. Pseudocode~\ref{alg:pcmci} summarizes the full 
PCMCI procedure implemented in this study.

\begin{algorithm}[!t]
\caption{PCMCI for multivariate time series causal discovery}
\label{alg:pcmci}
\begin{algorithmic}[1]

\State \textbf{Input:} Multivariate time series $\{\mathbf{X}_t\}_{t=1}^T$ with $d$ variables; maximum lag $\tau_{\max}$; thresholds $\alpha_{\mathrm{PC}}$ and $\alpha_{\mathrm{MCI}}$.
\State \textbf{Output:} Set of directed lagged edges $\mathcal{E}$.

\State Construct candidate parent sets:
\[
\mathcal{P}_0(X_t^{(i)}) =
\big\{ X_{t-\tau}^{(j)} \,\big|\, 1 \le j \le d,\; 1 \le \tau \le \tau_{\max} \big\}.
\]

\Statex
\State \textbf{Stage 1: PC condition selection}

\For{$i = 1$ to $d$}
    \State $\mathcal{P}(X_t^{(i)}) \gets \mathcal{P}_0(X_t^{(i)})$
    \State $k \gets 0$
    \While{conditioning sets of size $k$ exist}
        \ForAll{$X_{t-\tau}^{(j)} \in \mathcal{P}(X_t^{(i)})$}
            \State For all $\mathbf{Z} \subseteq \mathcal{P}(X_t^{(i)}) \setminus \{X_{t-\tau}^{(j)}\}$ with $|\mathbf{Z}| = k$, test:
            \[
            H_0:\;
            X_{t-\tau}^{(j)} \; \perp\!\!\!\perp \; X_t^{(i)}
            \;\big|\; \mathbf{Z}.
            \]
            \If{any $p > \alpha_{\mathrm{PC}}$}
                \State Remove $X_{t-\tau}^{(j)}$ from $\mathcal{P}(X_t^{(i)})$
            \EndIf
        \EndFor
        \State $k \gets k+1$
    \EndWhile
\EndFor

\Statex
\State \textbf{Stage 2: MCI testing}

\State $\mathcal{E} \gets \emptyset$

\For{$i = 1$ to $d$}
    \ForAll{$X_{t-\tau}^{(j)} \in \mathcal{P}(X_t^{(i)})$}
        \State Define the MCI conditioning set:
        \[
        \mathbf{Z}_{\mathrm{MCI}} =
        \mathcal{P}(X_t^{(i)}) \setminus \{ X_{t-\tau}^{(j)} \}.
        \]

        \State Test the dependence condition:
        \[
        X_{t-\tau}^{(j)} \; \not\!\perp\!\!\!\perp \; X_t^{(i)}
        \;\big|\; \mathbf{Z}_{\mathrm{MCI}}.
        \]

        \If{$p \le \alpha_{\mathrm{MCI}}$}
            \State Add edge $(X_{t-\tau}^{(j)} \to X_t^{(i)})$ to $\mathcal{E}$
        \EndIf
    \EndFor
\EndFor

\State \Return $\mathcal{E}$

\end{algorithmic}
\end{algorithm}

\subsubsection{Causal Effect Estimation}
\label{subsubsec:causal_effect_estimation}
In this section, we demonstrate how our proposed method quantifies causal
effects under linear structural dependencies. To estimate the causal
contribution of each marketing channel to conversions, we use a Structural
Causal Model (SCM) implemented with the \texttt{DoWhy} framework%
\footnote{\url{https://www.pywhy.org/dowhy}}. An SCM represents each variable
$X_i$ through a structural equation
\begin{equation}
    X_i = f_i\!\big( \mathcal{P}(X_i^{t}), N_i \big),
\end{equation}
where $\mathcal{P}(X_i^{t})$ denotes the causal parents of $X_i^{t}$ and $N_i$
is an exogenous noise term. After inferring the causal graph, each causal
mechanism $f_i$ is estimated directly from the observed data.

In causal inference, several estimands characterize the effect of an intervention,
including the Average Treatment Effect (ATE), the Conditional Average Treatment
Effect (CATE), the Individual Treatment Effect (ITE), the Average Treatment
Effect on the Treated (ATT), the Average Treatment Effect on the Controls (ATC),
and the Local Average Treatment Effect (LATE)
\cite{imbens2015causal,cheng2022evaluation}. Each estimand contrasts potential
outcomes under treatment $(T = 1)$ and control $(T = 0)$, but they differ in how
the averaging is performed and which subpopulations are targeted.

In our setting, each observation is generated by a structural causal model and
corresponds to a specific configuration of channel impressions and contextual
covariates. Therefore, the causal effect of a marketing channel on conversions
must be evaluated \emph{conditional} on these observed features. The CATE is
well suited for this purpose, as it measures how the expected conversion outcome
changes when a specific channel is intervened upon (e.g., setting its impression
level to zero) while holding all remaining covariates fixed.

To quantify the effect of removing a particular marketing channel, we express
the intervention using the do-operator. Let $X_j$ denote the activity of channel
$j$ (e.g., impressions), $Y_t$ the conversion outcome, and $Z_t$ the collection
of all other observed covariates at time $t$. We consider a binary intervention
in which the channel is either kept at its normalized active level ($X_j = 1$)
or entirely removed ($X_j = 0$). The CATE of channel $j$ at covariate profile
$z$ is then
\begin{equation}
    \Delta_j(z)
    =
    \mathbb{E}\!\left[ Y_t \,\middle|\, \mathrm{do}(X_j = 1),\, Z_t = z \right]
    -
    \mathbb{E}\!\left[ Y_t \,\middle|\, \mathrm{do}(X_j = 0),\, Z_t = z \right],
\end{equation}
and we approximate these conditional expectations by generating interventional samples from the fitted SCM while conditioning on the empirical distribution of $Z_t$.

Because these counterfactual expectations are not available in closed form, we estimate them using interventional Monte Carlo sampling. For each run $i$, we generate two sets of interventional samples conditioned on the observed covariates $Z_t = z$: one under the intervention $\mathrm{do}(X_j = 1)$ and one under $\mathrm{do}(X_j = 0)$. This yields empirical means $\overline{Y}_{\mathrm{do}(X_j = 1),\, i}$ and $\overline{Y}_{\mathrm{do}(X_j = 0),\, i}$. Over $n$ runs, the conditional causal effect of channel $j$ at covariate profile $z$ is then estimated by

\begin{equation}
\widehat{\Delta}_j(z)
= \frac{1}{n}
  \sum_{i=1}^{n}
  \left(
  \overline{Y}_{\mathrm{do}(X_j = 1),\, i}
  -
  \overline{Y}_{\mathrm{do}(X_j = 0),\, i}
  \right),
\end{equation}
which converges to the true CATE $\Delta_j(z)$ as $n$ increases.

Repeating this sampling procedure over n runs yields a stable estimate of the conditional causal effect $\widehat{\Delta}_j(z)$, representing the expected change in conversions when channel $j$ is switched from its active state $\mathrm{do}(X_j = 1)$ to the inactive state $\mathrm{do}(X_j = 0)$ at covariate profile $z$. This Monte Carlo procedure also provides uncertainty quantification without requiring parametric assumptions on the structural equations.

To express the causal effect on a comparable scale across channels, we compute a per-unit causal contribution by normalizing the estimated effect by the average activity level of channel $j$:
\begin{equation}
    \mathrm{ACE\ per\ unit}(j)
    =
    \frac{\widehat{\Delta}_j(z)}{\mathbb{E}[X_j]}.
\end{equation}
This quantity represents the incremental causal contribution of a single unit of channel activity (e.g., one impression) to expected conversions.

\subsection{Evaluation Metrics}
\label{subsec:evaluation-metrics}

\subsubsection{Causal DAG Discovery}
To quantify the DAG structures learned by our proposed causal-discovery algorithm, we adopt two complementary families of metrics: (i) \emph{edge‐based graph classification measures} to compare the number of edges that differ between two graphs \cite{cheng2022evaluation}, and (ii) \emph{graph-distance-based measures}, which evaluate how well the learned graph supports valid identifying formulas for effects, and. Below we go through each measure in detail.

The edge-based graph distance will be based on the intuition that directional adjacency relations can be treated as a binary classification problem. Therefore, a variety of metrics in classification tasks can be used such as AUC, True Positive Rate (TPR), False Positive Rate (FPR), and $F_{\beta}$. 

\paragraph{\textbf{True Positive Rate and False Positive Rate}} 
We consider a finite set of random variables \(X_{1}, \ldots, X_{D}\) with index set 
\(\mathbf{V} = \{1, \ldots, D\}\). A graph \(\mathcal{G} = (\mathbf{V}, \mathcal{E})\) consists 
of the node set \(\mathbf{V}\) and an edge set \(\mathcal{E} \subseteq \mathbf{V} \times \mathbf{V}\).

Let \(\mathcal{G}_{\text{true}} = (\mathbf{V}, \mathcal{E}_{\text{true}})\) denote the ground truth DAG, 
and let \(\mathcal{G}_{\text{pred}} = (\mathbf{V}, \mathcal{E}_{\text{pred}})\) denote the predicted DAG.

The true positive rate (TPR) is defined as the ratio of correctly recovered edges, meaning 
the edges that appear in both \(\mathcal{E}_{\text{pred}}\) and \(\mathcal{E}_{\text{true}}\), relative to 
the total number of edges in the ground truth graph. Formally,
\begin{equation}
    \text{TPR}
    = \frac{|\mathcal{E}_{\text{pred}} \cap \mathcal{E}_{\text{true}}|}{|\mathcal{E}_{\text{true}}|}.
\end{equation}

The false positive rate (FPR) measures the proportion of incorrectly added edges, meaning 
the edges that appear in \(\mathcal{E}_{\text{pred}}\) but not in \(\mathcal{E}_{\text{true}}\), relative 
to the number of edges that are not present in the ground truth graph. Formally,
\begin{equation}
    \text{FPR}
    = \frac{|\mathcal{E}_{\text{pred}} \setminus \mathcal{E}_{\text{true}}|}
           {|(\mathbf{V} \times \mathbf{V}) \setminus \mathcal{E}_{\text{true}}|},
\end{equation}
where \(\mathbf{V} \times \mathbf{V}\) denotes the set of all possible directed edges among the 
variables, excluding self loops if desired.

TPR and FPR are computed from the p-value matrices after applying a p\% (e.g., 5\%) significance threshold. TPR measures the fraction of true causal links that the method correctly identifies, while FPR measures the fraction of absent links that the method incorrectly labels as causal. Formally, TPR is the number of correctly detected links divided by the number of true links, and FPR is the number of falsely detected links divided by the number of absent links. Ideally, a causal discovery algorithm should recover true causal links with high sensitivity while keeping false positive (incorrect link) detections under control.

\paragraph{\textbf{Area Under the ROC Curve}} 
Area under ROC curve (AUC) is the area under the curve of recall versus FPR at different thresholds. The AUC quantifies the ability of the model to distinguish true causal links from non-links based on the score matrices returned by PCMCI. Each score matrix contains non-negative values, where a larger entry at position $(i, j)$ indicates greater confidence in a causal edge $i \rightarrow j$. The corresponding ground-truth label is binary (1 for a true causal link, 0 otherwise). Higher AUC values, with a maximum of 1, indicate stronger separation between true and false links.

\paragraph{\textbf{F-measure ($F_{\beta}$)}}
The F-measure, also computed on thresholded p-value matrices, evaluates the harmonic mean of precision and recall. It ranges from 0 (poor performance) to 1 (perfect precision and recall). In this study, we use $F_{0.5}$, which places greater emphasis on precision than recall, reducing the impact of false negatives. This makes $F_{0.5}$ particularly suitable when the cost of false positives is higher or when identifying spurious edges is more detrimental than missing weak ones.

Nevertheless, the number of differing edges between two graphs does not capture how the graphs differ with respect to the identifying formulas they imply for causal effects. Therefore, edge-based metrics cannot quantify whether a predicted graph preserves the causal semantics encoded by the true graph. In particular, two graphs may differ in several edges yet induce the same interventional distributions, while another pair may differ by only a few edges but imply fundamentally different causal effects. To evaluate causal fidelity rather than purely structural similarity, we additionally employ graph-distance-based measures that compare graphs in terms of the causal relationships they encode, as described in the next section.

\paragraph{\textbf{Structural Hamming Distance}} 

The Structural Hamming Distance (SHD) \citep{acid2003searching, tsamardinos2006max} is one of the most commonly used metrics for evaluating the accuracy of a learned causal graph. Intuitively, SHD counts the number of edge insertions, deletions, and orientation reversals required to transform the learned graph into the true graph. A low SHD value indicates a more accurate reconstruction of the true causal structure, whereas a higher SHD reflects greater structural deviation and thus poorer causal discovery performance.

Given a true causal graph $G_{\text{true}}$ and a predicted (learned) graph $G_{\text{pred}}$, both defined over the same vertex set $\mathbf{V}$, we define their adjacency matrices $A_{\text{true}}, A_{\text{pred}} \in \{0,1\}^{D \times D}$ such that $A[i,j] = 1$ if there is an edge $X_i \to X_j$.

The Hamming distance between binary vectors measures the number of directed edges that differ between the two adjacency matrices. Formally,

\begin{equation}
\begin{aligned}
\mathrm{SHD}(G_{\text{true}}, G_{\text{pred}})
    &= \sum_{i=1}^{D} \sum_{j=1}^{D}
        |A_{\text{true}}[i,j] - A_{\text{pred}}[i,j]| \\[4pt]
    &= \|A_{\text{true}} - A_{\text{pred}}\|_{1}.
\end{aligned}
\end{equation}

Alternatively, for each unordered pair of nodes $(i,j)$ with $i<j$, define
\begin{equation}
    \delta_{ij} =
    \begin{cases}
    1, & 
    \text{if } A_{\text{true}}[i,j] \neq A_{\text{pred}}[i,j] \\[2pt]
       & \text{or } A_{\text{true}}[j,i] \neq A_{\text{pred}}[j,i], \\[6pt]
    0, & \text{otherwise}.
    \end{cases}    
\end{equation}

Then,
\begin{equation}
    \mathrm{SHD}(G_{\text{true}}, G_{\text{pred}}) 
        = \sum_{1 \le i < j \le D} \delta_{ij}. 
\end{equation}

A normalized variant scales the distance by the number of possible unordered node pairs:
\begin{equation}
{n{SHD}}(G_{\text{true}}, G_{\text{pred}}) 
    = \frac{1}{\binom{D}{2}}
      \sum_{1 \le i < j \le D} \delta_{ij},
\end{equation}
so that ${n\mathrm{SHD}} \in [0,1]$. This normalization expresses the proportion of differing edges relative to the total possible pairwise relations.

While the SHD implementation is simple and efficient to compute, it only reflects local structural differences and does not account for causal equivalence or d-separation patterns. A very small structural change (e.g., flipping a single edge $A \rightarrow B \rightarrow C$  to $A \leftarrow B \rightarrow C$) can completely alter the causal implications of a graph even when SHD suggests that the two graphs are nearly identical. Consequently, SHD has been criticized for failing to capture the full causal implications of structural differences between graphs. This limitation has motivated the development of more causally informed distance measures, such as the Structural Intervention Distance (SID), which will be discussed below.

\paragraph{\textbf{Structural Intervention Distance}} 
The Structural Intervention Distance (SID) counts the number of ordered node pairs $(i,j)$
for which the interventional distributions implied by the true DAG and the learned DAG differ.
To understand SID, we first introduce the intervention distribution.

Given a directed acyclic graph (DAG) $\mathcal{G}$ over variables 
$\mathbf{X} = (X_1, \ldots, X_p)$ with joint distribution $\mathcal{L}(\mathbf{X})$,
the parents of a node $X_j$ are defined as
\begin{equation}
\mathrm{PA}^{\mathcal{G}}_j = \{\, X_i : (i,j) \in \mathcal{E} \,\},
\end{equation}
meaning that $X_i$ is a parent of $X_j$ whenever the edge $X_i \rightarrow X_j$ is present.

Pearl's do-operator formalises interventions by replacing the conditional distribution 
of the intervened node. Intervening on $X_j$ with a distribution $\tilde{p}(x_j)$ yields
\begin{equation}
p(x_1,\ldots,x_p \mid do(X_j = \tilde{p}(x_j)))
    = \prod_{i \neq j} p(x_i \mid x_{\mathrm{PA}_i}) 
       \cdot \tilde{p}(x_j),
\end{equation}
whenever $p(x_1,\ldots,x_p) > 0$, and zero otherwise.

A point intervention $do(X_j = \tilde{x}_j)$ is obtained by choosing 
$\tilde{p}(x_j) = \delta_{x_j,\tilde{x}_j}$:
\begin{equation}
p(x_1,\ldots,x_p \mid do(X_j = \tilde{x}_j))
    = \prod_{i \neq j} p(x_i \mid x_{\mathrm{PA}_i})
       \cdot \delta_{x_j,\tilde{x}_j}.
\end{equation}

Often we are interested only in the marginal effect of $X_j$ on another variable $X_i$.
If $X_i$ is not a parent of $X_j$, the parent adjustment formula 
(Theorem 3.2.2 in \cite{pearl2009causality}) gives:
\begin{equation}
p(x_i \mid do(X_j = \tilde{x}_j))
    = \sum_{x_{\mathrm{PA}_j}}
        p(x_i \mid \tilde{x}_j, x_{\mathrm{PA}_i})
        \, p(x_{\mathrm{PA}_j}).
\end{equation}

SID compares a true graph $\mathcal{G}_{\text{true}}$ with an estimated graph 
$\mathcal{G}_{\text{pred}}$ by counting the number of node pairs $(i,j)$ 
for which the interventional distributions differ:
\begin{equation}
\begin{aligned}
\mathrm{SID}(\mathcal{G}_{\text{true}}, \mathcal{G}_{\text{pred}})
   = \Big|\, \big\{ (i,j) :\;
        & p_{\mathcal{G}_{\text{pred}}}(X_j \mid do(X_i))  \\
        & \neq\;
          p_{\mathcal{G}_{\text{true}}}(X_j \mid do(X_i))
     \big\} \,\Big|.
\end{aligned}
\end{equation}

A pair $(i,j)$ is called \emph{good} if the intervention distribution obtained from the learned graph matches that from the true graph for all observational distributions $\mathcal{L}(\mathbf{X})$. Otherwise, the pair contributes one unit to the SID score. In this way, SID evaluates errors in causal ordering, whereas SHD counts only incorrect edges.

For a graph with $V$ nodes, the total number of ordered node pairs 
$(i,j)$ with $i \neq j$ is
\begin{equation}
N_{\text{pairs}} = V(V-1).
\end{equation}

The normalized Structural Intervention Distance is defined as
\begin{equation}
\mathrm{nSID}(\mathcal{G}_{\text{true}}, \mathcal{G}_{\text{pred}})
    = \frac{
        \mathrm{SID}(\mathcal{G}_{\text{true}}, \mathcal{G}_{\text{pred}})
      }{
        V(V-1)
      }.
\end{equation}

Therefore, nSID measures the proportion of causal effects that the learned graph gets wrong. Values close to 0 indicate high causal accuracy, whereas values close to 1 indicate poor causal estimation.

\subsubsection{Causal Effect Estimation}

After estimating the CATE for each marketing channel (as shown in Section~\ref{subsubsec:causal_effect_estimation}), we evaluate the accuracy of the causal-driven attribution model using three complementary metrics.

\paragraph{\textbf{Relative Root Mean Square Error}}
The relative root mean square error (RRMSE) measures the scale-normalized deviation between the estimated causal effects $\hat{\tau}_i$ and the ground-truth effects
$\tau_i$:
\begin{equation}
    \text{RRMSE}
    = \frac{\sqrt{\frac{1}{n}\sum_{i=1}^{n}(\hat{\tau}_i - \tau_i)^2}}
           {\frac{1}{n}\sum_{i=1}^{n}|\tau_i|},
\end{equation}
where $n$ is the number of channels. This metric captures overall estimation error relative to the magnitude of true effects.

The interpretation of the RRMSE depends entirely on the context, domain, and units of the outcome variable. Different research areas adopt different thresholds for what constitutes an acceptable level of normalized error. For example, in the energy and agricultural modeling literature, model accuracy is often considered \emph{excellent} when $\text{RRMSE} < 10\%$, \emph{good} when $10\% \leq \text{RRMSE} < 20\%$, \emph{fair} for $20\% \leq \text{RRMSE} < 30\%$, and \emph{poor} when $\text{RRMSE} \ge 30\%$ \cite{jamieson1991test,despotovic2016evaluation}. These thresholds illustrate that RRMSE benchmarks are not universal but rather depend on the application context and the expected signal-to-noise ratio in the data.

\paragraph{\textbf{Mean Absolute Percentage Error}}
The mean absolute percentage error (MAPE) expresses the average proportional deviation:
\begin{equation}
    \text{MAPE}
    = \frac{100\%}{n}\sum_{i=1}^{n}
      \left|\frac{\hat{\tau}_i - \tau_i}{\tau_i}\right|.
\end{equation}
RRMSE and MAPE capture complementary aspects of estimation accuracy. RRMSE gives more weight to larger errors because it is based on \textit{squared deviations} that are normalized by the scale of the true effects, making it sensitive to large misestimations. In contrast, MAPE provides an interpretable percentage-based measure of accuracy by averaging \textit{absolute proportional errors}. For instance, a MAPE of $5\%$ indicates that the estimated causal effects deviate from the true effects by only $5\%$ on average. Using both metrics allows us to assess estimation quality in terms of both scale-normalized sensitivity to large errors (RRMSE) and average relative deviation (MAPE).

\paragraph{\textbf{Spearman Rank Correlation}}

To evaluate whether the attribution model preserves the relative importance of marketing channels, we compute the Spearman rank correlation between the ground truth causal effects and the estimated causal effects obtained from the model \cite{cornell2024rank,arshad2024predictive}.

Suppose $C$ denotes the list of marketing channels sorted (either ascending or descending) according to their true causal effects (ground-truth ranking), and $C_p$ denotes the list of the same channels sorted by their estimated causal effects from the attribution model. For each channel $x$, we define its rank in the true and predicted lists as
\begin{equation}
    \begin{aligned}
        r_{C}(x)   &= \text{position of channel } x \text{ in } C, \\
        r_{C_p}(x) &= \text{position of channel } x \text{ in } C_p.
    \end{aligned}
\end{equation}

From these ranks, we construct two ranked vectors:
\begin{equation}
    \begin{aligned}
    R_{C} = \big( r_{C}(x_1),\, r_{C}(x_2),\, \ldots,\, r_{C}(x_n) \big), \\
    R_{C_p} = \big( r_{C_p}(x_1),\, r_{C_p}(x_2),\, \ldots,\, r_{C_p}(x_n) \big), 
    \end{aligned}
\end{equation}
where $n$ is the total number of marketing channels.

The Spearman rank correlation coefficient $\rho$ between the two ranked lists is defined as
\begin{equation}
    \rho = 1 - \frac{6 \sum_{i=1}^{n} d_i^2}{n(n^2 - 1)},
\end{equation}
where $d_i = r_{C}(x_i) - r_{C_p}(x_i)$ is the difference in rank for the $i^{th}$ channel, and $n$ is the total number of marketing channels.

Spearman's $\rho$ takes values between $-1$ and $1$. A value of $\rho = 1$ indicates perfect agreement, meaning the model recovers the exact ordering of true causal contributions across marketing channels. A value of $\rho = -1$ means the estimated ranking is the exact reverse of the true ranking. A value of $\rho = 0$ indicates no monotonic relationship between the estimated and true rankings. In our study, a high Spearman correlation signifies that the model correctly identifies which channels are more or less influential, even if the estimated magnitudes differ. This property is particularly important for resource allocation and budget optimization, where the correct ordering of channel importance often matters more than precise effect magnitudes.

\section{Results}
This study uses \textit{Tigramite}\footnote{\url{https://jakobrunge.github.io/tigramite/}}, 
an open-source Python package for reconstructing graphical models 
(conditional independence graphs) from multivariate time series 
based on the PCMCI framework, and \textit{DoWhy}\footnote{\url{https://github.com/py-why/dowhy}}
 for estimating causal effects.

The algorithm proceeds in two steps. First, a PC step screens candidate parent sets using a permissive threshold ($\alpha_{\mathrm{PC}} = 0.2$) to favour recall and preserve potentially relevant dependencies. Second, a Momentary Conditional Independence (MCI) test re-evaluates each retained link at a stricter significance level ($\alpha_{\mathrm{MCI}} = 0.05$), ensuring robustness and suppressing spurious associations. This two-step procedure balances false negatives and false positives, yielding stable edge sets across repeated runs.

To ensure a DAG causal structure, bidirectional links are pruned by retaining the edge whose time-lagged effect has the greater magnitude, and any remaining cycles are resolved using a priority rule that preserves edges directed into the main outcome variable (e.g., conversions). This enforces causal directionality toward the dependent variable of interest.

\subsection{Optimal Time-delay Causality}
Nevertheless, a major challenge when using PCMCI is specifying the appropriate time lag needed to reliably estimate causal links. We do not know in advance which time lag best reflects the underlying causal relationships in the simulated data. Therefore, an empirical assessment is necessary to identify which lag values allow PCMCI to most reliably recover the underlying causal structure in our simulated environment.

\begin{figure*}[ht!]
    \centering

    \subfigure[]{
        \includegraphics[width=0.48\textwidth]{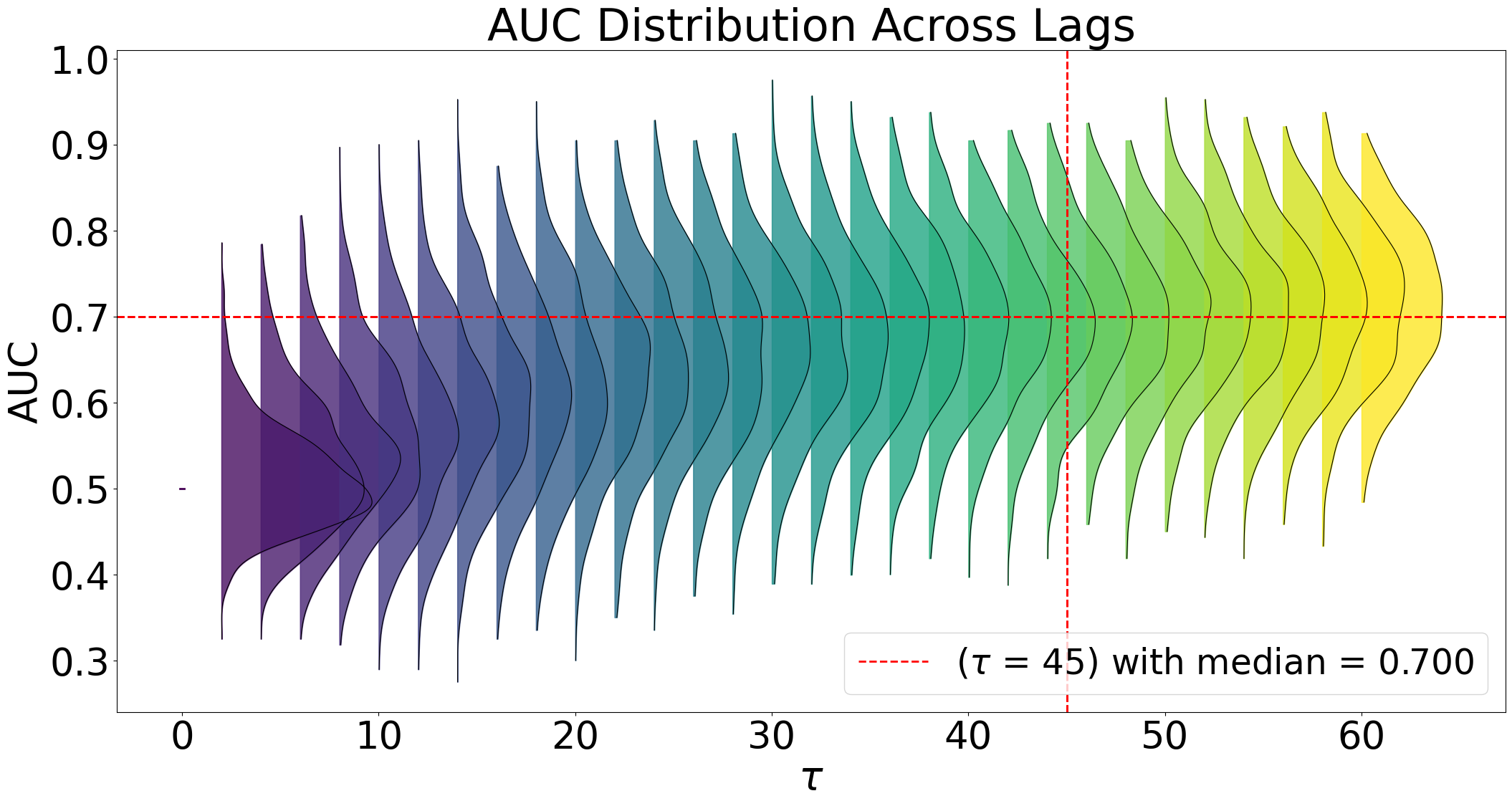}
    }
    \hfill
    \subfigure[]{
        \includegraphics[width=0.48\textwidth]{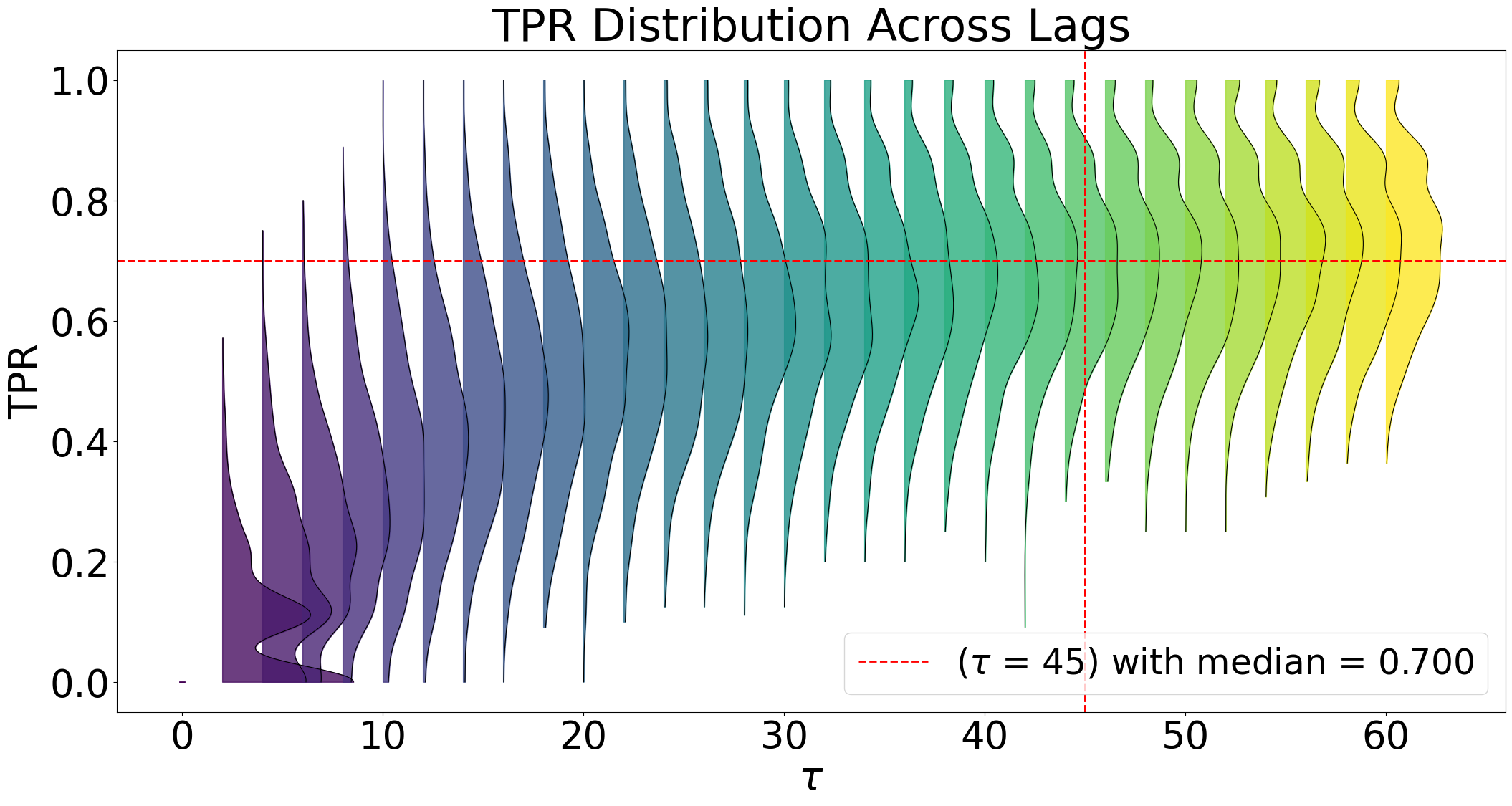}
    }

    \subfigure[]{
        \includegraphics[width=0.48\textwidth]{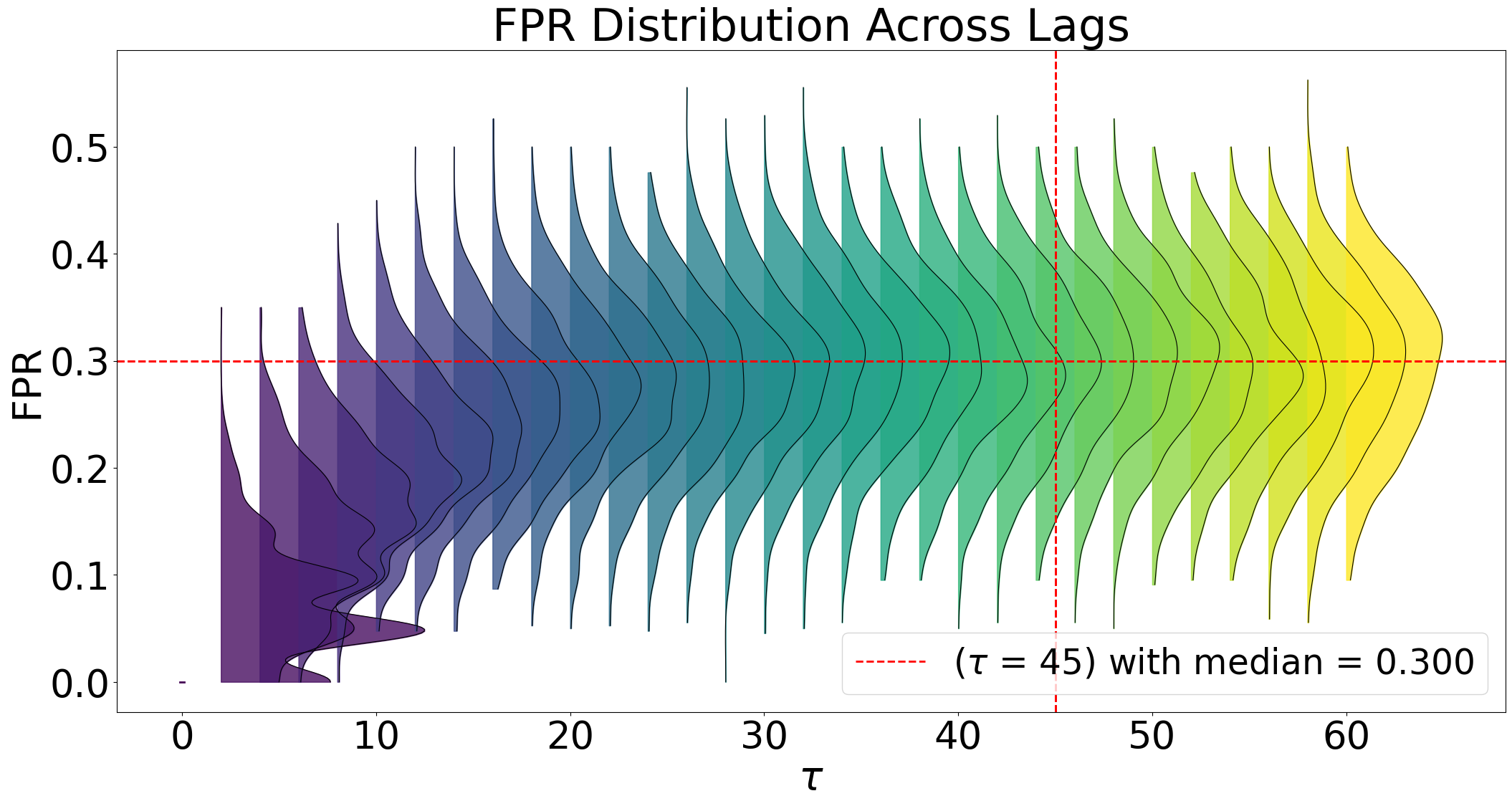}
    }
    \hfill
    \subfigure[]{
        \includegraphics[width=0.48\textwidth]{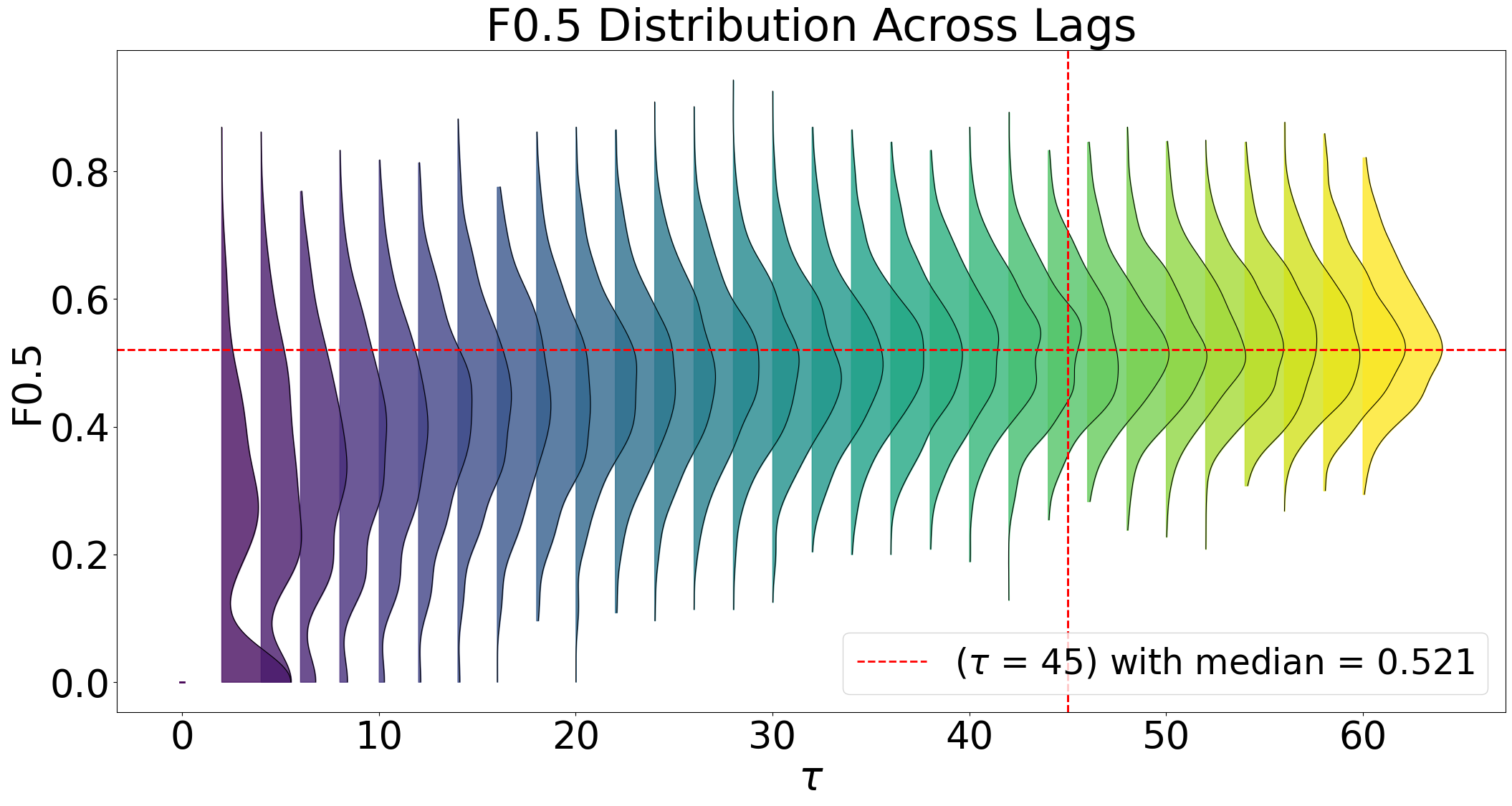}
    }

    \subfigure[]{
        \includegraphics[width=0.48\textwidth]{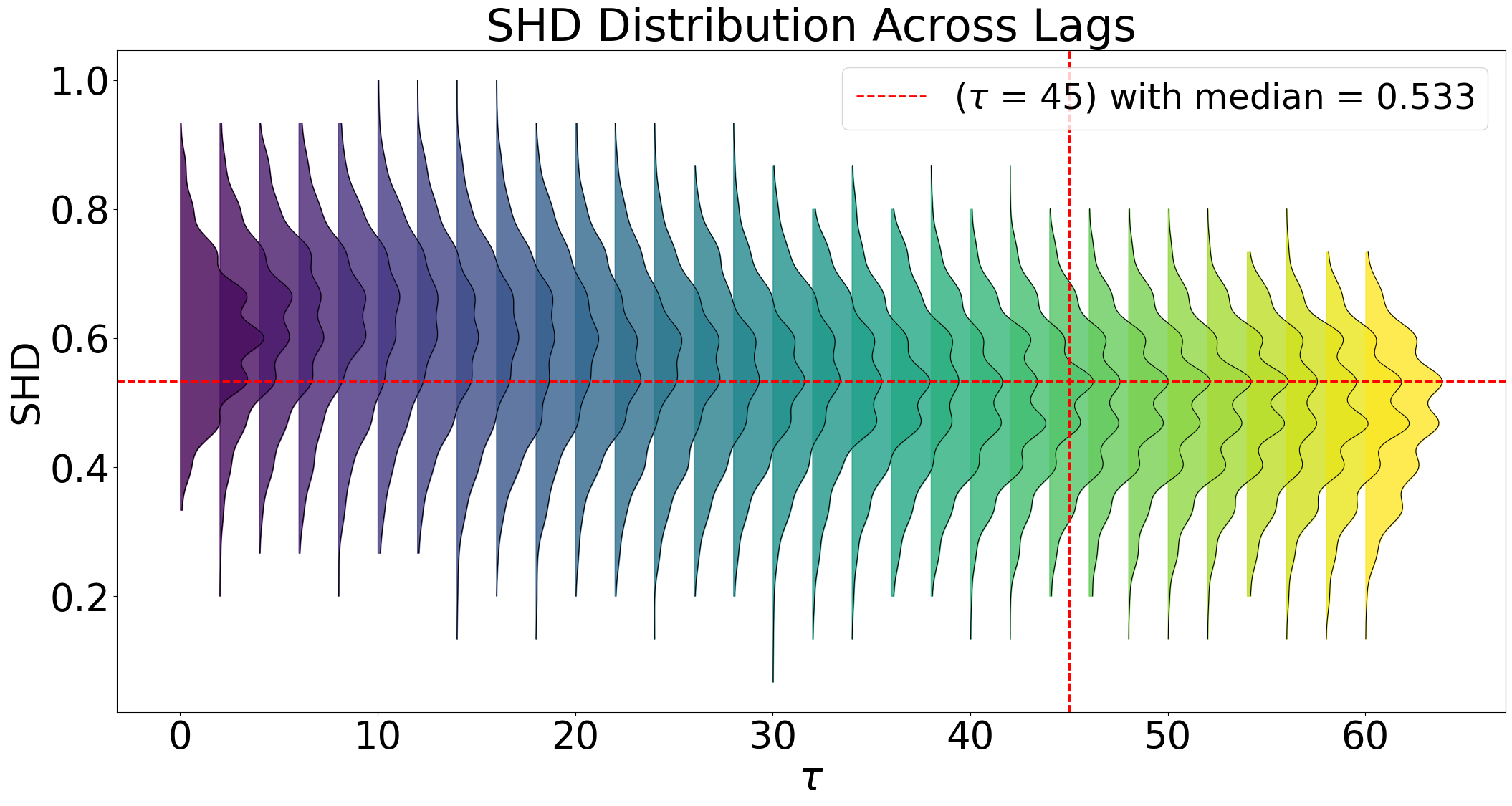}
    }
    \hfill
    \subfigure[]{
        \includegraphics[width=0.48\textwidth]{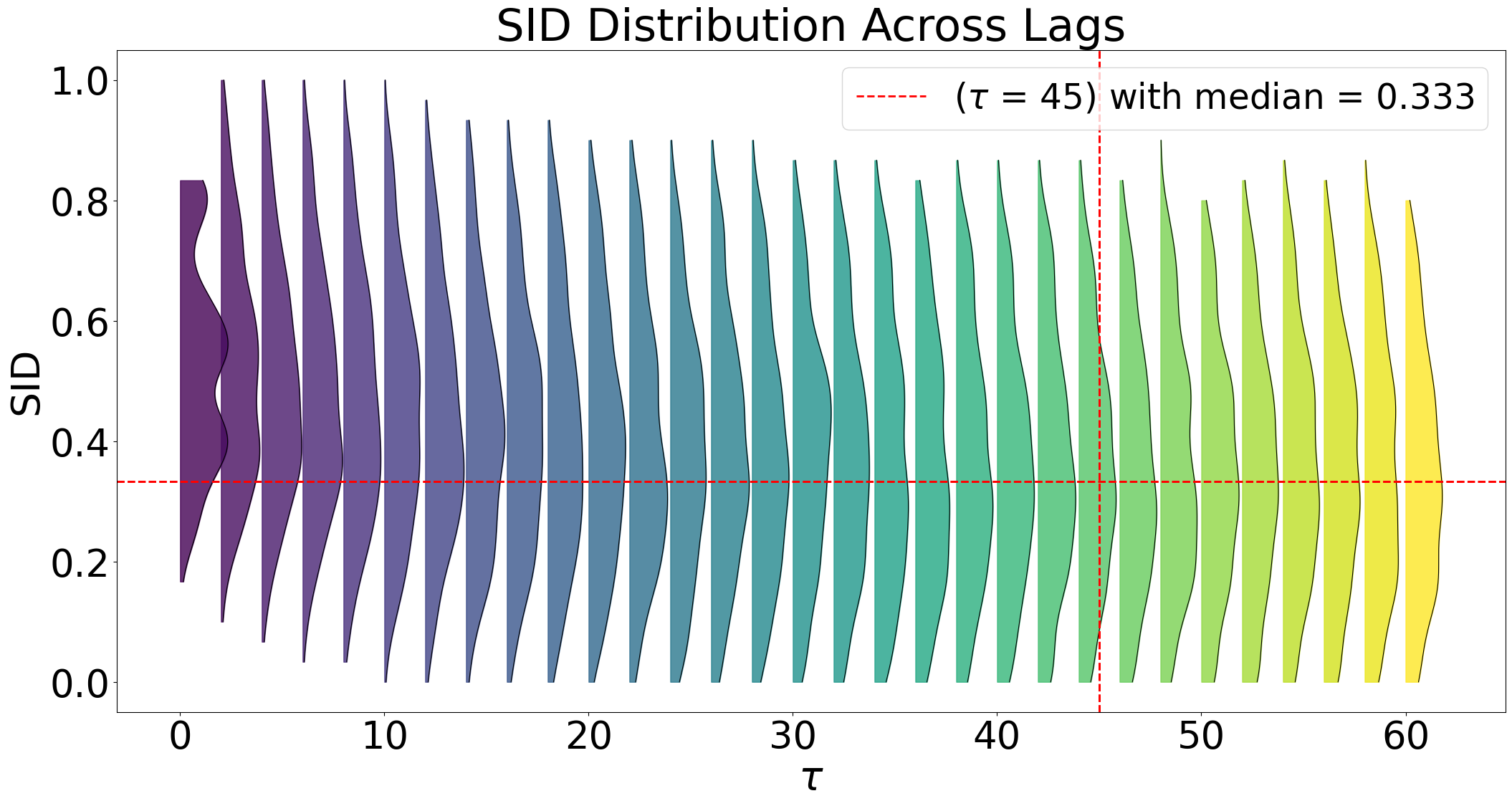}
    }

    \caption{Kernel density estimates (KDEs) of six evaluation metrics, including AUC, TPR, FPR, F0.5, SHD, and SID across lag values ($\tau$). Each subplot shows how the empirical distribution of the metric varies as $\tau$ increases from 0 to 60, using 1000 simulation seeds for each $\tau$. The viridis color gradient is used to visually distinguish KDE ridges at different lag values; color does not encode density or any statistical quantity. Instead, the horizontal width of each ridge reflects the density and spread of metric values at a given lag. The red dashed vertical line marks $\tau = 45$, and the red dashed horizontal line denotes the corresponding median metric value at that lag.}
    \label{fig:tau_results}
\end{figure*}

To this end, we explore temporal lags up to $\tau_{\max} = 60$ and evaluate
model performance across $1,000$ simulated datasets. For each lag value
$\tau$, we compute several predictive and structural metrics, including AUC, TPR, FPR, F0.5, SHD, and SID, and summarize their empirical distributions using kernel density estimates (KDEs), as visualized in Fig.~\ref{fig:tau_results}. Each ridge in the figure corresponds to $1,000$ simulations at a fixed lag and illustrates how both the central tendency and variability of each metric evolve as $\tau$ increases.

Across all six metrics, a consistent pattern emerges: performance steadily
improves with increasing lag, reaches its optimum around $\tau = 45$, and then
either plateaus or slightly deteriorates. At $\tau = 45$, the KDE ridges indicate three desirable properties: \emph{(i)}~the best concentration of high-performing metric values, \emph{(ii)}~the lowest variability across seeds, and \emph{(iii)}~a balanced trade-off between predictive accuracy (AUC, TPR, F0.5) and structural recovery (SHD, SID). In addition, the FPR remains well controlled around this lag, meaning it does not increase substantially despite gains in sensitivity (TPR). Such behaviour is consistent
with a saturation effect, where additional lag increases no longer introduce new false positives because most meaningful dependencies have already been captured. This confluence of high performance and low uncertainty identifies $\tau = 45$ as the optimal lag value $\tau_{max}$.

In conclusion, this result indicates that the simulated system exhibits a characteristic time delay of approximately 45 time steps between channel impressions and observed conversions. At this lag, PCMCI most effectively captures the temporal propagation of causal influence across variables. For this reason, all subsequent causal attribution experiments adopt $\tau_{max} = 45$ as the default time-delay configuration.

\subsection{Causal-Driven Attribution Analysis}
\subsubsection{Illustrative Example}
\begin{figure*}[!htpb]
    \subfigure[True Causal DAG]{
        \includegraphics[width=0.48\textwidth,height=0.28\textheight]{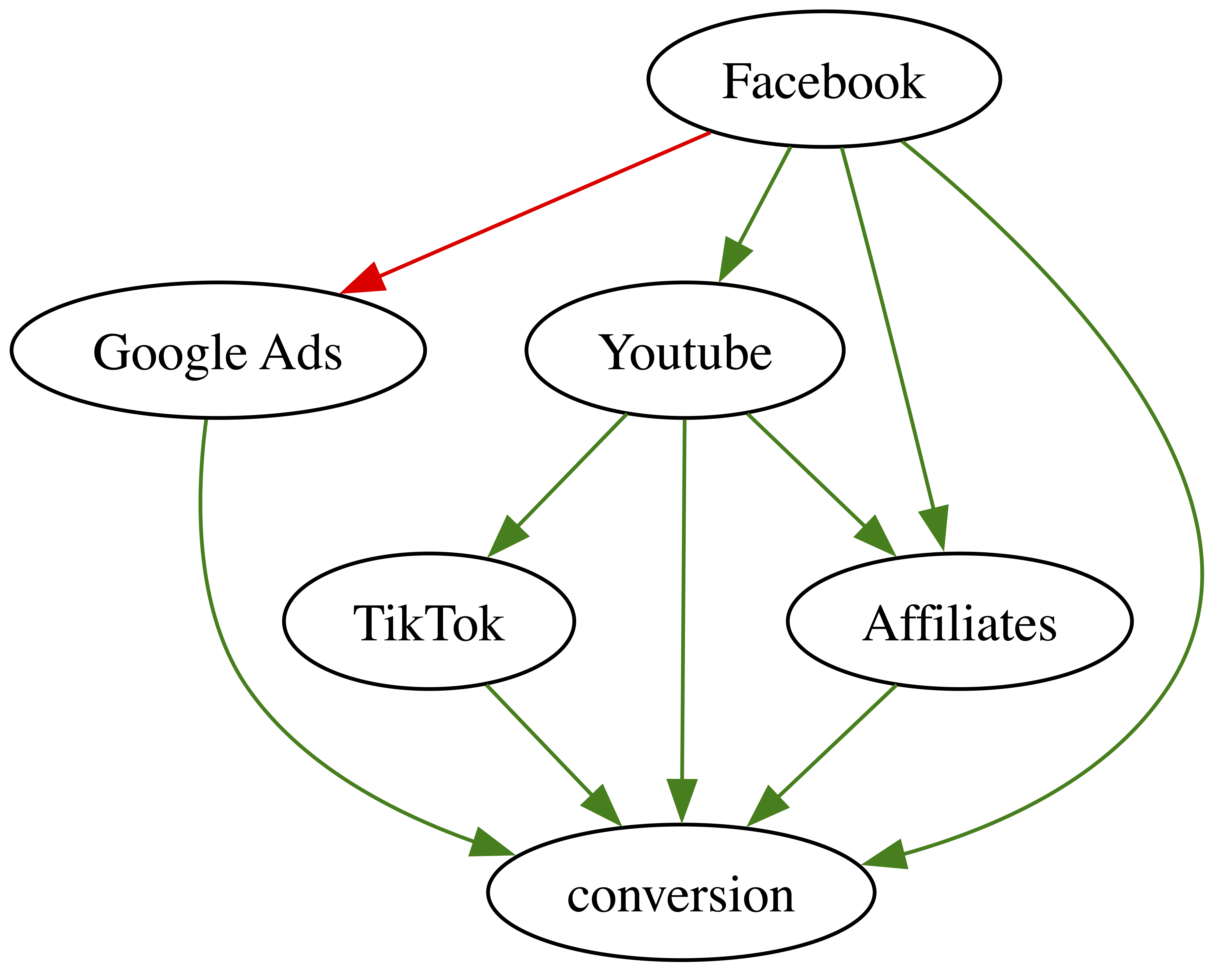}
    }
    \hspace{0.03\textwidth}
    \subfigure[Predicted Causal DAG]{
        \includegraphics[width=0.47\textwidth,height=0.32\textheight]{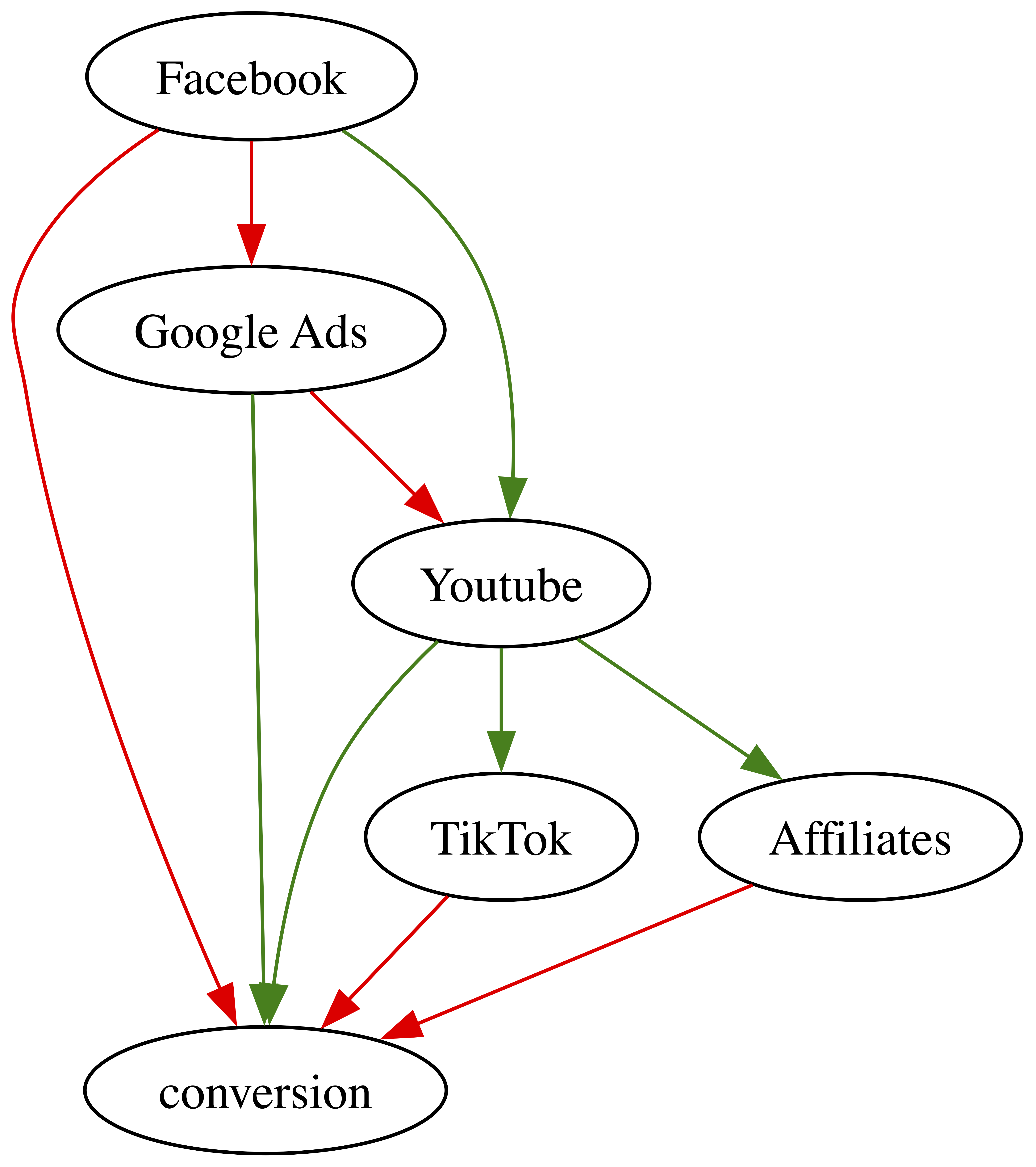}
    }

    \caption{Comparison between the ground-truth DAG (a) and the DAG predicted using PCMCI (b).}
    \label{fig:dag_sample_comparison}
\end{figure*}

\paragraph{\textbf{Causal DAG Discovery}}
Fig.~\ref{fig:dag_sample_comparison} compares the ground-truth causal graph (left) with the graph predicted by the PCMCI algorithm (right). The true DAG is generated from our simulated structural model (with \textit{seed} value 155 in this illustration), where each directed edge corresponds to a known causal relationship and all effects on the target variable, \textit{conversion}. Of course, the choice of seed is arbitrary, and any random seed would produce a valid instance of the simulated structural model. Overall, our estimator successfully recovers the major components of the underlying structure.

According to graph-distance–based metrics, the comparison determined an SHD of 0.133 and an SID of 0.10, indicating strong agreement with the ground-truth DAG in both its structural configuration and its implied interventional behaviour. These low values suggest that only a small number of edge discrepancies remain and that the learned graph preserves most of the true causal dependencies.

From a classification perspective, we also obtained a strong performance. The TPR is 0.90, which means that 90\% of actual causal edges were correctly identified. The FPR is relatively low at 0.05, indicating that the predicted graph is well calibrated in its edge decisions. Furthermore, the AUC is 0.925, showing that our method achieves a high level of discriminative power when ranking edges by their likelihood of being causal. The $F_{0.5}$ score is 0.90, which confirms that the majority of the identified edges are accurate and not attributable to false detections.

Although our causal graph discovery approach recovers much of the true structure, it also introduces minor errors. For example, it fails to detect a true edge from \textit{Facebook} $\rightarrow$ \textit{Affiliates}, and it incorrectly introduces spurious edges such as \textit{Google Ads} $\rightarrow$ \textit{YouTube}. Additionally, the learned graph contains certain sign errors, where the direction of causal influence is correct but the estimated effect is negative rather than positive. To address these issues, a post-estimation refinement step will be applied to improve the accuracy of the predicted DAG.

\paragraph{\textbf{Causal Effect Estimation}}
To assess the performance of our causal effect estimator, we first evaluate it under ideal conditions by providing the true causal DAG as input. Since the simulated data are generated from this ground-truth structure, the estimated causal effects should closely match the true effects. This step verifies that the estimator is correctly specified and capable of recovering the underlying causal relationships when the structure is known. We then apply the estimation using the DAG learned by our discovery approach. To obtain robust estimates of causal effects, we follow the procedure described in Section~\ref{subsubsec:causal_effect_estimation} and compute CATE using a Monte Carlo approach. The estimation process is repeated 5,000 times under independent stochastic draws, and the resulting estimates are averaged to reduce variance.

\begin{figure*}[!htpb]
    \subfigure[CDA result on True Causal Graph]{
        \includegraphics[width=0.48\textwidth,height=0.34\textheight]{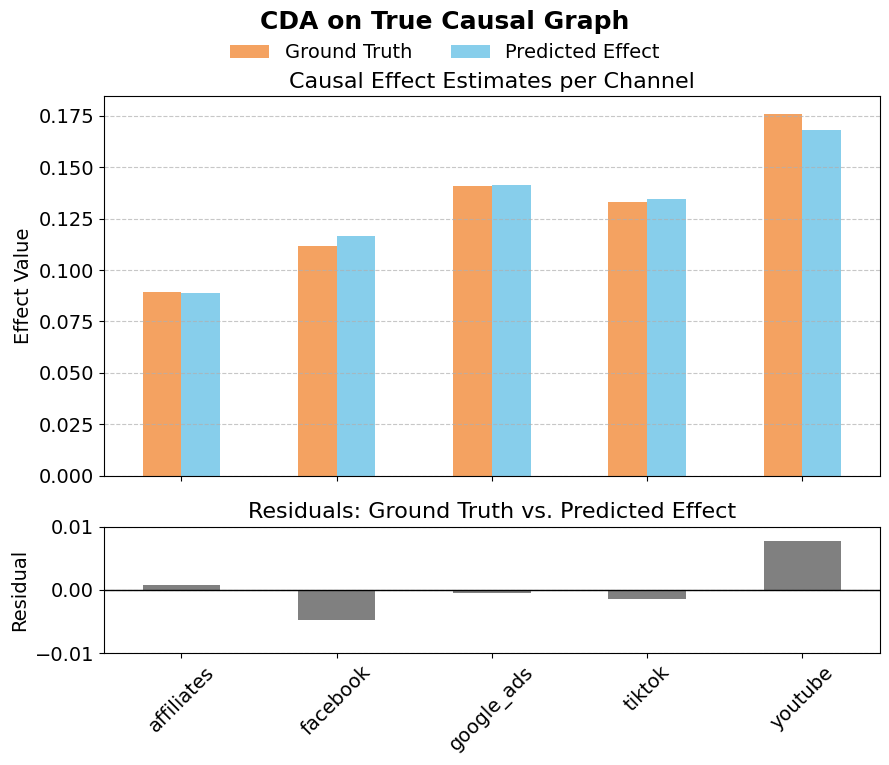}
    }
    \hspace{0.02\textwidth}
    \subfigure[CDA result on Predicted Causal Graph]{
        \includegraphics[width=0.48\textwidth,height=0.34\textheight]{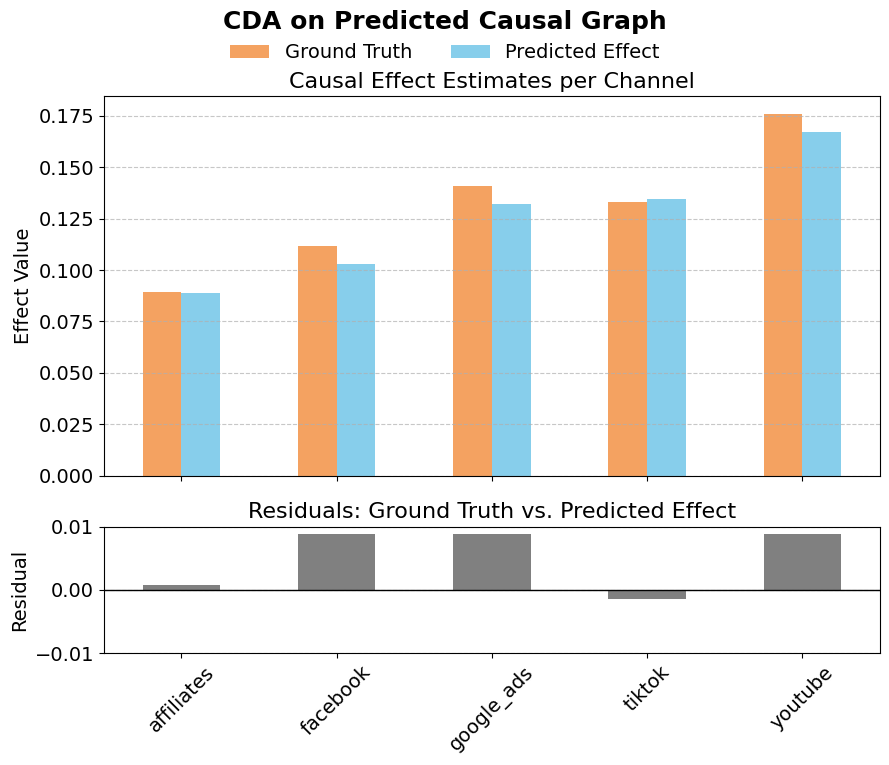}
    }

    \caption{Comparison of causal effect estimates and residuals for (a) the predicted causal graph and (b) the simulated ground-truth causal graph, compared to the true causal effects. Each panel shows the estimated causal effects per marketing channel and the residual differences between estimated and true values.}
    \label{fig:cda_sample_comparison}
\end{figure*}

It is important to note that small deviations are expected even when using the true graph. Each channel inherently contains noise, which causes estimation error to accumulate along causal pathways with multiple parent nodes. For example, in the graph shown in Fig.~\ref{fig:dag_sample_comparison} (a), both \textit{Facebook} and \textit{YouTube} have several outgoing connections, which increases noise propagation when the estimator calibrates their effects. As a result, channels with direct causal links such as \textit{Affiliates}, \textit{Google Ads}, and \textit{TikTok} show nearly perfect recovery, whereas \textit{Facebook} and \textit{YouTube} display slightly higher residual error. This expectation is indeed reflected in the results shown in Fig.~\ref{fig:cda_sample_comparison} (a).

When the estimator is applied to the predicted DAG (as depicted in Fig.~\ref{fig:cda_sample_comparison} (b), the overall accuracy maintains a high level of accuracy, with effect estimates that closely reflect the true underlying causal effects across channels. As expected, the residuals increase slightly relative to the true-graph case, particularly for channels whose incoming or outgoing edges were misidentified in the learned structure. Nevertheless, the deviations are small, implying that the predicted graph is sufficiently accurate to support reliable causal effect estimation.

These observations are quantitatively supported by the results presented in Table~\ref{tab:sample_estimation_metrics}. Using the true graph yields very low estimation error, with RRMSE of 3.16\% and MAPE of 2.17\%, along with perfect rank preservation as indicated by a Spearman correlation of 1.00. When the predicted graph is used, error measures increase moderately to RRMSE of 5.27\% and MAPE of 4.20\%, while rank correlation remains high at 0.90. This indicates that although structural imperfections in the learned graph introduce some additional estimation error, the overall magnitude of causal effects and their relative ordering across channels are still captured with high fidelity.

In summary, the results indicate that the estimator performs effectively whether it receives the true DAG or the predicted DAG as input, confirming that the workflow is both reliable and robust. However, this evaluation is based on only a single synthetic instance. To obtain a more reliable and statistically grounded assessment, the next section presents a broader simulation study. Using the same evaluation metrics, this extended analysis provides deeper insight into the performance of our causal-driven attribution framework.

\begin{table}[ht]
\centering
\caption{Comparison of causal effect estimation accuracy using the true causal graph versus the predicted graph. Lower RRMSE and MAPE indicate better estimation accuracy, while higher Spearman correlation reflects better rank preservation.}
\begin{tabular}{lccc}
\toprule
\textbf{DAG Input} & \textbf{RRMSE (\%)} & \textbf{MAPE (\%)} & \textbf{Spearman Corr} \\
\midrule
True Graph      & 3.16 & 2.17 & 1.00 \\
Predicted Graph & 5.27 & 4.20 & 0.90 \\
\bottomrule
\end{tabular}
\label{tab:sample_estimation_metrics}
\end{table}

\subsubsection{Aggregate Simulation Results}
Table~\ref{tab:multi_simulation_results} summarizes the estimation performance across 1,000 simulations, generated by sampling 200 random DAGs for each depth level from 2 to 6 layers (corresponding to our setup with five input channels and one target). This balanced design ensures that each layer of structural complexity contributes equally to the overall results. Several insights emerge regarding the behaviour and robustness of the causal effect estimator as DAG depth increases and the underlying causal architecture becomes more complex.

\begin{table*}[t]
\centering
\caption{Estimation performance across DAG depths. Values report mean $\pm$ standard deviation over 200 simulations per depth.}
\label{tab:multi_simulation_results}
\resizebox{\textwidth}{!}{
\begin{tabular}{lccccccc}
\toprule
\textbf{Input} & \textbf{Metric} & \textbf{2 Layers} & \textbf{3 Layers} & \textbf{4 Layers} & \textbf{5 Layers} & \textbf{6 Layers} & \textbf{All} \\
\midrule

\multirow{3}{*}{\textbf{True DAG}}
& \textbf{RRMSE (\%)} 
& $0.79 \pm 0.44$ 
& $7.71 \pm 7.36$ 
& $10.48 \pm 9.38$ 
& $13.56 \pm 10.81$ 
& $14.96 \pm 12.67$
& $9.50 \pm 10.44$ \\

& \textbf{MAPE (\%)} 
& $0.69 \pm 0.38$ 
& $4.43 \pm 4.08$ 
& $6.32 \pm 5.80$ 
& $9.00 \pm 7.77$ 
& $10.18 \pm 9.00$
& $6.12 \pm 7.05$ \\

& \textbf{Spearman $\rho$} 
& $0.98 \pm 0.05$ 
& $0.92 \pm 0.16$ 
& $0.87 \pm 0.24$ 
& $0.84 \pm 0.22$ 
& $0.84 \pm 0.23$
& $0.89 \pm 0.20$ \\

\midrule

\multirow{3}{*}{\textbf{Predicted DAG}}
& \textbf{RRMSE (\%)} 
& $12.92 \pm 10.20$ 
& $20.65 \pm 12.97$ 
& $25.50 \pm 13.72$ 
& $29.34 \pm 15.37$ 
& $32.76 \pm 15.43$
& $24.23 \pm 15.31$ \\

& \textbf{MAPE (\%)} 
& $9.31 \pm 7.38$ 
& $15.97 \pm 10.72$ 
& $22.34 \pm 26.91$ 
& $24.52 \pm 14.13$ 
& $26.85 \pm 14.07$
& $19.80 \pm 17.26$ \\

& \textbf{Spearman $\rho$} 
& $0.78 \pm 0.29$ 
& $0.63 \pm 0.38$ 
& $0.50 \pm 0.44$ 
& $0.40 \pm 0.47$ 
& $0.37 \pm 0.49$
& $0.53 \pm 0.45$ \\
\bottomrule
\end{tabular}
}
\end{table*}

\paragraph{\textbf{Performance when using the true DAG}}
Across different DAG depths, the estimator demonstrates strong accuracy when provided with the true causal DAG. Both RRMSE and MAPE remain low for shallower graphs with two to three layers and increase gradually as depth grows, reflecting the inherent difficulty of recovering causal effects in structures with more parent–child dependencies and greater noise accumulation. Even in the deepest configurations with five to six layers, the estimation error remains moderate relative to the scale of the simulated effects. The Spearman rank correlation also remains consistently high, ranging from 0.84 to 0.98, indicating that the estimator reliably preserves the relative ordering of channel importance even as absolute estimation error increases with graph complexity.

The aggregated results across all depths provide further confirmation of the estimator's robustness. With an overall RRMSE of 9.50\%, MAPE of 6.12\%, and Spearman correlation of 0.89, the estimator recovers the true causal effects with high fidelity on average across a broad range of DAG structures. These metrics demonstrate that the proposed approach maintains strong performance under the ideal condition of perfect structural knowledge, establishing a performance benchmark against which results using the predicted graph can be evaluated. The consistent preservation of rank ordering is particularly noteworthy from a practical standpoint, as marketing decision-makers often prioritize understanding the relative importance of channels over obtaining precise point estimates of individual effects.

\paragraph{\textbf{Performance when using the predicted DAG}}
When the estimator relies on the predicted graph rather than the true DAG, estimation error increases, as expected. This decline in performance is statistically significant in deeper DAGs (five to six layers), where inaccuracies in the predicted structure intensify the inherent complexity of the task. For instance, the RRMSE grows sharply from 12.92\% at depth 2 to 32.76\% at depth 6, and the MAPE shows a similar pattern, rising from 9.31\% to 26.85\%. The Spearman correlation also drops steadily as the number of layers increases, falling from 0.78 at two layers to 0.37 at six layers. This indicates that inaccuracies in the predicted structure make it harder for the model to correctly rank the importance of different channels. 

Even so, the overall performance across all simulations indicates that the estimator remains informative. With an RRMSE of 24.23\% and MAPE of 19.80\%, the estimated causal effects differ from the true magnitudes by roughly 20 to 25 percent on average. While this represents a non-trivial loss of numerical precision compared to using the true graph, it still falls within a practically acceptable range for most real-world attribution scenarios, where measurement noise and structural uncertainty are unavoidable. The moderate Spearman $\rho$ of 0.53 further demonstrates that the estimator retains meaningful discriminatory power, preserving the ability to distinguish high-impact channels from lower-impact ones, even though the ranking is no longer perfectly reliable.

Taken together, these results demonstrate that the proposed causal-driven attribution approach remains useful under realistic conditions, providing meaningful insight into relative channel contributions despite the presence of substantial structural noise in the learned DAGs.

\paragraph{\textbf{Effect of DAG depth}}
Both the true-DAG and predicted-DAG results reveal a clear and consistent pattern: deeper graphs with a greater number of causal layers systematically reduce estimation accuracy. This decline is intuitive, as increasing depth introduces more intermediate nodes and pathways through which random noise can propagate, thereby amplifying variance in the estimated effects. Additionally, deeper structures create more opportunities for small errors in the estimated relationships to accumulate and magnify as they propagate downstream through the causal chain. Whether these errors stem from incorrect edge magnitudes or from the presence or absence of connections in the learned graph, their cumulative impact grows substantially with each additional layer.

The impact of depth becomes particularly pronounced when the estimator relies on the predicted DAG rather than the true structure. In this scenario, inaccuracies in the learned graph interact with the inherent complexity of deeper causal architectures, resulting in compounding estimation errors that degrade both numerical precision and rank-ordering performance. This heightened deterioration underscores the sensitivity of causal effect estimation to structural misspecification, especially in settings where causal pathways span multiple layers or involve densely connected nodes.

Overall, the results demonstrate that while the estimator maintains robust performance for shallower graphs, deeper causal architectures pose substantially greater challenges. The degradation manifests both in terms of absolute numerical accuracy and in the model's ability to preserve the correct relative ordering of channel effects. These findings have practical implications for marketing attribution. In contexts where channel interactions are known to involve multiple layers of indirect influence, practitioners should anticipate greater uncertainty in effect estimates and may need to place greater emphasis on directional insights and relative rankings rather than precise point estimates. The results also highlight the value of pursuing accurate structural discovery, as even modest improvements in the learned graph can yield meaningful gains in downstream estimation quality when dealing with complex causal systems.

\section{Discussion and conclusion}
Now more than ever, businesses rely on data-driven decision making to allocate marketing resources and maximize return on investment. However, the growing integration of AI into data driven technologies in digital advertising, such as recommendation systems \cite{jabbar2020real} and digital platforms \cite{purmonen2023b2b}, has made consumer journeys increasingly complex, creating multi touchpoint pathways that span numerous channels \cite{micheaux2019customer, roy2025business, hardcastle2025understanding}. Simultaneously, the rise of big data has introduced challenges related to missing and uncertain information, complicating the interpretation of marketing signals. Furthermore, stringent privacy regulations (as discussed in Section~\ref{subsec:regulation_challenges}), together with platform level restrictions such as the deprecation of third party cookies, have sharply reduced access to the user-level data required by traditional attribution models. Consequently, conventional attribution methods that rely on tracking individual user journeys across touchpoints have become increasingly impractical and, in many cases, impossible to implement. This confluence of factors creates an urgent need for new attribution approaches that can operate effectively in privacy-constrained environments while still capturing the complex interdependencies that characterize modern marketing ecosystems.

To the best of our knowledge this is the first ever causal-driven attribution utilising impression data without path information. When introducing a new attribution model, it is essential to consider the standards and challenges it must address to be practically viable. The first and most fundamental issue concerns the type of data it requires. As discussed in the literature review, numerous attribution methodologies exist, and many provide reliable results that businesses can trust. However, a major limitation of existing attribution approaches is their dependence on user-level path data. These models can only operate when detailed sequences of individual user interactions across touchpoints are available. If a business does not collect or have access to such data, it cannot apply these methods regardless of its analytical capabilities or resources. Therefore, our causal-driven attribution solved this challenges by using alternative data structures, in particular aggregated data that does not require tracking individual user journeys while still representing the relationships between marketing channels and conversions.

Second, our approach uses impressions as independent variables, representing a necessary evolution beyond path-based attribution models that depend primarily on clicks or visits. As privacy regulations restrict user level tracking, organisations lose the ability to validate platform reported click counts, which makes the collection of accurate and reliable click records progressively more challenging. This shift aligns with a broader industry recognition that impressions capture valuable brand exposure that may not generate immediate interactions but can significantly influence downstream consumer behaviour over time. As discussed in Section~\ref{subsec:regulation_challenges}, we have concluded that exposure effects often precede and facilitate subsequent conversions, even in the absence of direct click engagement. Therefore, impression-based models not only address the practical reality of diminished click tracking but also provide a more comprehensive view of marketing influence by accounting for both direct response mechanisms and longer-term awareness-building effects that traditional click-centric approaches overlook.

Third, the trustworthiness of attribution methods is grounded in their ability to uncover meaningful causal relationships among marketing channels rather than simple correlations \cite{hair2021data}. Marketing practitioners must understand which channels truly drive outcomes, not merely which ones exhibit statistical associations with conversions. Our causal driven attribution model addresses this requirement by estimating channel effects within a structured causal framework. Although many researchers have developed reliable attribution approaches, most of these methods rely on user path data (as discussed in Section~\ref{subsec:causal_imperative}). While the use of aggregated data does prevent the identification of causal effects at the level of individual user journeys, this limitation is not problematic in practice. Businesses make strategic budget decisions at the population level, where the primary concern is the overall influence of channels rather than the behaviour of specific individuals. Without a causal understanding of these effects, firms risk allocating resources to channels that appear effective but do not actually contribute to performance. For this reason, an attribution model that operates on aggregated data must still deliver credible causal inference. Our approach is designed to meet this requirement and support actionable marketing decisions.

Finally, because our approach does not require Personally Identifiable Information, it can be used in environments where tracking individual users is restricted by privacy or security rules. This makes it especially valuable for highly regulated industries such as pharmaceuticals, telecommunications, healthcare, and education, where strict data protection laws often make traditional tracking methods impractical or illegal \cite{onifade2021advances,seth2025innovations}. Beyond these regulated sectors, the framework offers a practical solution for any organization that needs effective marketing attribution while avoiding the substantial infrastructure costs associated with collecting, storing, and securing user-level tracking data. By operating on aggregated data, our approach reduces both financial barriers and compliance burdens, thereby making advanced attribution analysis accessible to organisations of varied sizes and capacities.

\subsection{Theoretical implications}
This paper contributes to the marketing and causal inference literature by proposing a novel causal-driven attribution framework designed to operate exclusively on aggregated marketing data, specifically impression-level time series, without relying on path-level or user-level tracking. This addresses a critical and growing need for privacy-preserving, scalable, and interpretable attribution models in marketing analytics, particularly in regulatory environments where access to granular, identifiable data is increasingly restricted or prohibited.

First of all, although Structural Equation Modelling is traditionally employed for path analysis and hypothesis testing \cite{stein2011structural}, its generative nature makes it well suited for simulating synthetic datasets with controlled causal structures. In this study, we leverage SEM as a mechanism to generate data that follow predefined structural relationships among marketing channels and conversion outcomes. By encoding causal pathways, functional forms, and noise processes directly into the model, SEM enables us to simulate realistic datasets that mimic the multivariate dependencies observed in marketing environments. While no simulation framework can perfectly reproduce real behavioral data, SEM provides a coherent and interpretable approximation of the underlying data-generating process through its regression-based formulation of structural relations \cite{iacobucci2009everything}. These modeling assumptions offer a transparent and theoretically grounded foundation for constructing synthetic data, enabling controlled experimentation while preserving essential features of causal structure, noise propagation, and channel interactions.

Our causal-driven attribution framework reconceptualizes attribution as a causal inference problem rather than a correlational one, aligning with insights from prior studies \cite{smith2003parachute,hair2021data}. It contributes to attribution theory by integrating causal discovery and causal estimation within a unified analytical pipeline. In the discovery stage, the framework employs the PCMCI algorithm to identify temporal conditional independencies and recover the directional structure among marketing channels using only impression-level time series, establishing a causal graph without requiring user-level tracking data. In the estimation stage, the inferred structure is used to compute Conditional Average Treatment Effects that quantify both direct and indirect channel contributions to conversions. A Monte Carlo procedure is applied to obtain stable effect estimates under uncertainty. Together, these stages offer a transparent and rigorous approach to attribution by linking structural discovery with causal effect quantification. Across multiple simulated settings, the framework achieves strong empirical performance, recovering true causal effects with an average RRMSE of 9.50\% and Spearman correlation of 0.89 when the true graph is provided, and maintaining reasonable accuracy under the predicted graph with an average RRMSE of 24.23\% and Spearman correlation of 0.53. This integration extends existing attribution theory by demonstrating how causal inference can be achieved from aggregated data while preserving interpretability and robustness.

A third contribution in theoretical implications of this study is the development of a comprehensive benchmarking framework for evaluating attribution models under controlled causal conditions. Since real world marketing data rarely reveal the true causal relationships among channels, empirical validation of attribution models is typically constrained or indirect. By constructing a simulation environment grounded in structural equation modelling, predefined causal graphs, and systematically varied graph layers, our approach generates datasets where the true channel effects are known by design. This enables robust, repeatable comparisons of attribution performance across a wide range of structural complexities and data generating processes. The benchmark not only facilitates the assessment of our own framework but also offers a foundation that future research can use to evaluate alternative attribution methods under transparent and reproducible conditions. As a result, it contributes to the field’s methodological foundations by establishing establishing a standard approach for testing causal attribution models when the true causal structure cannot be directly observed.

\subsection{Managerial implications}
Our causal-driven attribution model carries several practical implications for marketing managers, analysts, and decision-makers who operate in increasingly complex and privacy-constrained environments. First, managers no longer need to rely on user-level data or cookies to make data-driven decisions. Our approach applies only on aggregated impression-level data, making it highly applicable in industries where access to granular information is restricted by privacy regulations such as Apple's iOS14, GDPR, CCPA or technological changes such as cookie deprecation (as detailed in Section~\ref{subsec:regulation_challenges}. This opens the door for more organizations that can adopt evidence-based attribution strategies, including those that lack advanced data infrastructure.

A further managerial insight from this study concerns the ability of the causal-driven attribution framework to capture the overall marketing funnel. Many traditional attribution models, including machine learning and deep learning approaches, emphasize immediate conversion signals. They typically use user-path data to predict whether a specific sequence of interactions will lead to a conversion and then derive channel attribution by tracing model weights or feature contributions. Such approaches overlook the broader dynamics of how channels interact and influence one another throughout the customer journey. In contrast, our framework provides stakeholders with visibility into how digital channels work together across the funnel, revealing both direct and indirect effects. This holistic perspective enables decision-makers to understand the collaborative role of channels, which is essential for strategic planning and long-term business growth \cite{liu2022channel}.

An important managerial contribution is related to marketing budget allocation. With global advertising spending projected to exceed \$1.17 trillion in 2025 \cite{StatistaAdvertisingWorldwide}, effective budget allocation has become a critical managerial priority. Existing literature presents several approaches to budget allocation \cite{zhao2019unified,fischer2011practice}; however, most offer only partial visibility into the unique contributions of each channel within the overall marketing system. By estimating both the direct and indirect influence of each channel, our framework offers managers clearer and more actionable guidance for budget allocation. In addition, the ability to identify channels that exert negative or counterproductive influence enables firms to avoid wasteful spending and reallocate resources toward more effective activities.

Based on these insights, managers should reconsider how they evaluate performance metrics and KPIs within their marketing mix. Our method provides a more comprehensive perspective on channel influence, enabling a refinement of existing measurement practices. This expanded perspective can help managers rebalance their media portfolios by identifying undervalued yet effective touchpoints that may not be highlighted by traditional models. The simplicity and scalability of the framework make it suitable for both early stage firms with limited analytical capacity and large organizations overseeing complex multichannel campaigns. By reducing data requirements while enhancing the analytical richness, our approach helps democratize marketing measurement and supports more informed strategic decision making across diverse organizational contexts.

\subsection{Limitations and directions for future research}

This study has certain limitations that suggest promising directions for future work. We begin with the limitations of our causal discovery method. The model relies only on observational data and does not incorporate interventional evidence, which is known to enhance discovery performance \cite{silva2016observational}. Additionally, as noted by \citet{runge2019inferring}, causal inference is grounded in a set of underlying assumptions. Our approach assumes causal sufficiency, which means that all relevant variables are included and that no unobserved factors influence more than one channel. It also relies on the causal Markov condition, which states that each variable is independent of its non-effects given its direct causes. Finally, it assumes faithfulness, which requires that observed conditional independence relationships arise from genuine structural independence without spurious cancellations. Further research should explore the use of interventional data to strengthen causal identification and reduce reliance on strong structural assumptions. Researchers should also enhance the simulation framework by incorporating more realistic temporal patterns and behavioural dynamics that better represent actual marketing environments.

Second, a wide range of causal discovery methodologies has been developed, as documented in the comprehensive review by \citet{niu2024comprehensive}. Determining a suitable algorithm requires alignment between the analytical goals and the assumptions that can plausibly be applied to the data generating process. In this study, PCMCI was chosen because it is well suited to impression level time series and can accommodate temporal dependence structures that are common in marketing contexts. Nevertheless, other discovery techniques may achieve stronger performance under different modelling assumptions or where the data generating process follows structural patterns not well captured by PCMCI. The present work therefore provides a benchmark against which future research may evaluate and refine other causal discovery approaches, with the potential to enhance the accuracy and robustness of causal driven attribution models.

A further set of limitations arises from the causal estimation stage, which relies on observational data and therefore cannot fully eliminate the challenges associated with non-experimental inference. As \citet{li2014attributing} note, models estimated with non-experimental data are inherently vulnerable to endogeneity and omitted variable bias. Although our framework accounts for temporal ordering and conditional relationships, unobserved heterogeneity or strategic behaviour, such as selective targeting or endogenous channel activation, may still distort the estimated causal effects. These challenges are amplified in marketing settings, where channels often exhibit interdependent spillover and carryover dynamics that are difficult to separate without controlled interventions \cite{jin2017bayesian}. Future research could extend the present model by explicitly incorporating spillover and carryover mechanisms to better capture the interconnected nature of marketing activities.

Moreover, as \citet{ferketich1984residual} discuss, residual analysis in causal models often reveals violations of model assumptions, such as misspecification of functional forms, nonlinearity, and correlated errors across equations. These violations can undermine causal validity, particularly when feedback mechanisms or dynamic dependencies exist but are not explicitly modelled. In our framework, residual dependencies or systematic patterns may indicate modelled confounding or non-stationarity, which could lead to overestimation or underestimation of true causal effects. Future work should therefore explore robustness checks, alternative identification strategies, and the use of experimental or quasi-experimental designs to validate the causal pathways uncovered through our discovery and analysis stages.

As mentioned in Section~\ref{subsec:data-generation}, despite our efforts to identify a dataset suitable for comparison between our causal-driven attribution and traditional multi-touch attribution models, no such dataset was found. This is primarily because publicly available marketing datasets do not include the true causal effects or the underlying causal relationships among channels, both of which are necessary for a valid benchmark. We therefore constructed a synthetic dataset designed to preserve the statistical properties of real-world data \cite{raghunathan2021synthetic}. Prior studies \cite{yao2022causalmta,zhao2019unified} have likewise relied on synthetic data to evaluate attribution methodologies. We encourage future researchers to compare our CDA with ground truth derived from actual marketing data.

\section{Competing interests}
No competing interest is declared.

\section{Authors’ contributions}

G.F. was responsible for the ideation of the model, and G.F. initialized the results using the causal exploration algorithm. G.F. wrote the first, second, and third drafts of the publication. B.Q. provided valuable feedback on the algorithmic process and offered recommendations. B.Q. suggested the use of a synthetic dataset and created the process. B.Q. updated the model process, updated the results section, and provided feedback on model evaluation. B.Q. also wrote the model evaluation section. D.L. provided feedback on the readability and flow of the article. A.W. provided feedback on the method and evaluation process and suggested the initial model evaluation approach. A.J. provided feedback on the introduction and literature review and suggested relevant journals.

\section{Acknowledgments}
The authors thank the anonymous reviewers for their valuable suggestions.

\section{Code availability}
All scripts supporting the algorithm and data simulation will be made available upon publication. If you would like to access them beforehand, please contact the corresponding author.

\bibliographystyle{unsrtnat}
\bibliography{references}  






\end{document}